\documentclass[lettersize,journal]{IEEEtran}
\usepackage{amsmath,amsfonts}
\usepackage[linesnumbered,ruled,vlined]{algorithm2e}
\usepackage{array}
\usepackage[caption=false,font=normalsize,labelfont=sf,textfont=sf]{subfig}
\usepackage{textcomp}
\usepackage{stfloats}
\usepackage{url}
\usepackage{verbatim}
\usepackage{graphicx}
\usepackage{cite}

\usepackage{multirow}
\usepackage{booktabs}
\usepackage{hyperref}
\usepackage{amssymb}
\usepackage{subfloat}
\usepackage{subcaption} 

\usepackage{soul}
\soulregister\cite7
\soulregister\citep7 
\soulregister\citet7
\soulregister\ref7 
\soulregister\pageref7

\usepackage[table]{xcolor}
\usepackage{colortbl}
\usepackage{bm}

\begin{document}

\title{SAG-Agent: Enabling Long-Horizon Reasoning in Strategy Games via Dynamic Knowledge Graphs}

\author{Chenwei Tang, Lin Long, Xinyu Liu, Jingyu Xing, Zizhou Wang, Joey Tianyi Zhou,~\IEEEmembership{Senior Member,~IEEE,} \\ Jiawei Du*, Liangli Zhen*,~\IEEEmembership{Senior Member,~IEEE,}  Jiancheng Lv~\IEEEmembership{Senior Member,~IEEE}
\thanks{*: Co-corresponding Author.}
\thanks{This work is supported by National Major Scientific Instruments and Equipments Development Project of National Natural Science Foundation of China (No. 62427820) and Science Fund for Creative Research Groups of Sichuan Province Natural Science Foundation (No. 2024NSFTD0035).}
\thanks{C. Tang, L. Long, X. Liu, J. Xing, and J. Lv are with the College of Computer Science, Sichuan University, and Engineering Research Center of Machine Learning and Industry Intelligence, Ministry of Education, Chengdu 610065, PR China (e-mail: \{tangchenwei, lvjiancheng\}@scu.edu.cn, \{longlin1, xinyuliu0804, xingjingyuok\}@stu.scu.edu.cn).}
\thanks{Z. Wang and L. Zhen are with the Institute of High Performance Computing, Agency for Science, Technology and Research (A*STAR), Singapore 138632, Singapore (e-mail: wang\_zizhou@a-star.edu.sg, llzhen@outlook.com).}      
\thanks{J. Zhou and J. Du are with the Institute of High Performance Computing, and Centre for Frontier AI Research, Agency for Science, Technology and Research (A*STAR), Singapore 138632, Singapore (e-mail: zhouty@a-star.edu.sg, dujiawei@u.nus.edu).} 
\thanks{Manuscript received April 19, 2021; revised August 16, 2021.}}
\markboth{Journal of \LaTeX\ Class Files,~Vol.~14, No.~8, August~2021}%
{Shell \MakeLowercase{\textit{et al.}}: A Sample Article Using IEEEtran.cls for IEEE Journals}


\maketitle
\begin{abstract}
Most commodity software lacks accessible Application Programming Interfaces (APIs), requiring autonomous agents to interact solely through pixel-based Graphical User Interfaces (GUIs). In this API-free setting, large language model (LLM)-based agents face severe efficiency bottlenecks: limited to local visual experiences, they make myopic decisions and rely on inefficient trial-and-error, hindering both skill acquisition and long-horizon planning. To overcome these limitations, we propose \textbf{SAG-Agent}, an experience-driven learning framework that structures an agent's raw pixel-level interactions into a persistent State-Action Graph (SAG). SAG-Agent mitigates inefficient exploration by topologically linking functionally similar but visually distinct GUI states, constructing a rich neighborhood of experience that enables the agent to generalize from a diverse set of historical strategies. To facilitate long-horizon reasoning, we design a novel hybrid intrinsic reward mechanism based on the graph topology, combining a state-value reward for exploiting known high-value pathways with a novelty reward that encourages targeted exploration. This approach decouples strategic planning from pure discovery, allowing the agent to effectively value setup actions with delayed gratification. We evaluate SAG-Agent in two complex, open-ended GUI-based decision-making environments (\textit{Civilization V} and \textit{Slay the Spire}), demonstrating significant improvements in exploration efficiency and strategic depth over the state-of-the-art methods.

\end{abstract}


\section{Introduction}
\IEEEPARstart{T}{he} emergence of Large Language Models (LLMs) has accelerated a new generation of agentic Artificial Intelligence (AI) systems \cite{qin2025ui} capable of tackling complex tasks across a diverse range of domains, including web browsing \cite{zheng2023synapse}, operating mobile and desktop applications \cite{zhang2025ufo2}, crafting and exploration in virtual worlds \cite{wang2023describe}, and even robotics scenarios \cite{brohan2023can}. Initially, the dominant paradigm relied on agents based on predefined Application Programming Interfaces (APIs), where human designers decompose high-level goals into structured workflows and tools \cite{yao2023react}. While this approach ensures reliability on well-defined benchmarks \cite{xie2024osworld}, its dependence on task-specific APIs fundamentally limits the agent's adaptability and scalability in open-ended environments. Bridging this gap, recent progress in Vision Language Models (VLMs) \cite{zhou2022learning} has enabled Graphical User Interfaces (GUIs)-based agents \cite{zhang2025api}. These agents, such as UFO \cite{zhang2024ufo} and CogAgent \cite{hong2024cogagent}, interact not only through APIs but also by observing and manipulating GUIs in a human-like manner. By integrating visual understanding with reasoning, they provide more general automated control and a richer, more intuitive mode of interaction. The pursuit of ultimate generality and human-like autonomy takes this progression a final step further, leading to the API-free GUI-based agents, as exemplified by CRADLE \cite{tan2025cradle} and Bottom-Up Agent \cite{du2025rethinking}. This methodology envisions agents that operate exclusively through a universal, human-style interface—using only screen pixels as input and keyboard/mouse actions as output. Such agents have demonstrated the remarkable ability to autonomously acquire skills from scratch in complex games and diverse software applications without any privileged access.
\begin{figure}[t]
\centering
\includegraphics[width=\linewidth]{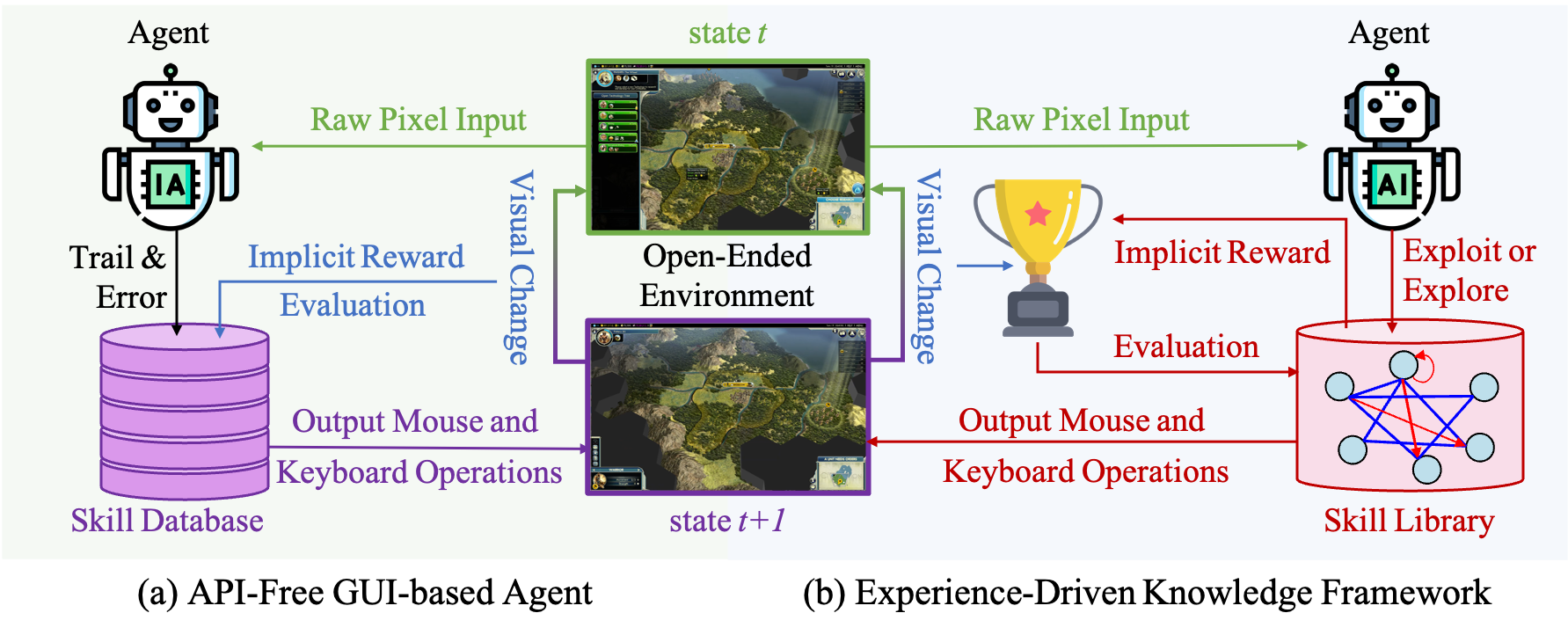}
\caption{From Local Memory to Global Strategy. (a) A conventional API-free GUI-based agent relies on a simple skill database, where experiences are isolated. Its decision-making is a reactive trial-and-error process guided by myopic rewards, leading to inefficient exploration. (b) Our agent leverages a structured knowledge graph as its skill library. This graph connects experiences, enabling a strategic \textit{exploit or explore} mechanism, and the learning of agent is guided by an advanced implicit reward that captures long-term value, accelerating knowledge transfer and fostering strategic planning.}
\label{fig:motivation}
\end{figure}

Despite these significant advances, as shown in Fig. \ref{fig:motivation} (a), the API-free GUI-based agents face two fundamental challenges \cite{du2025rethinking, tan2025cradle} that obstruct their path toward Artificial General Intelligence (AGI) \cite{morris2024position}: (i) \textit{\textbf{Inefficient Exploration}}: Without task-specific priors or APIs, skills must be discovered and validated purely through exhaustive trial-and-reasoning. This leads to a sample inefficiency so severe that agents typically require 2-2.5 times more environment interactions to match the progression of prior-assisted baselines. (ii) \textit{\textbf{Limited Long-Horizon Strategic Reasoning}}: The prevailing reliance on myopic reward signals, e.g., visual changes, fails to incentivize the multi-step plans essential for sophisticated gameplay. This inability to value setup actions with delayed gratification severely limits strategic depth. Overcoming these challenges requires methods that accumulate reusable abstractions from experience and reward formulations that capture long-term value, thus enabling robust planning, transfer, and generalization across diverse tasks.

To overcome these limitations, we introduce \textbf{SAG-Agent}, a novel experience-driven learning framework designed to transform an agent's pixel-based GUI interactions into structured, actionable knowledge. Central to our approach is a persistent, cross-episode State-Action Graph (SAG) that serves as the agent's long-term memory. This graph organizes raw visual observations into state nodes and models acquired skills as edges between them, thereby converting unstructured pixel-level experience into a coherent network for strategic decision-making in API-free environments. To directly combat \textit{\textbf{inefficient exploration}}, SAG-Agent connects functionally analogous yet visually distinct GUI states through similarity edges, forming a rich \textit{neighborhood of experience}. This enables the agent's reasoning module to query entire functional neighborhoods, generalizing from diverse historically successful strategies rather than relying on isolated, myopic data points. To address \textit{\textbf{limited long-horizon reasoning}}, we leverage the graph topology to design a novel hybrid intrinsic reward mechanism. This system combines a \textit{state value reward} (quantifying strategic potential based on outgoing connections) with a \textit{novelty reward} for environmental discovery. This potential-based formulation is crucial for incentivizing setup actions that yield delayed gratification, enabling robust long-term planning in open-ended environments. Our contribution is three-fold:
\begin{itemize}
    \item We propose \textbf{SAG-Agent}, a framework that structures raw GUI interaction experience into a persistent SAG, introducing a \textit{neighborhood of experience} to combat inefficient exploration by enabling generalization across functionally similar but visually distinct states.    
    \item We design a hybrid reward mechanism derived from the SAG's topology that addresses limited long-horizon reasoning. By combining a potential-based \textit{state value reward} for strategic exploitation with a \textit{novelty reward} for targeted exploration, our approach effectively incentivizes multi-step plans with delayed gratification.
    \item We conduct extensive experiments in two complex, open-ended environments (\textit{Civilization V} and \textit{Slay the Spire}), demonstrating that \textbf{SAG-Agent} significantly enhances exploration efficiency and strategic depth compared to state-of-the-art baselines.
\end{itemize}

\section{Related Work}
\label{sec:relatedwork}
\textbf{LLM-based Agents.} The application of LLM-based agents to complex, multi-step tasks has yielded remarkable achievements across diverse domains \cite{xi2025rise}, including web browsing \cite{gu2024your}, software operation \cite{jin2024llms}, and robotics \cite{firoozi2025foundation}. Video games pose a particularly demanding testbed, requiring precise low-level control alongside high-level planning, abstraction, and adaptation \cite{li2025comprehensive}. Recent advances have shown strong performance in environments like Minecraft \cite{li2025optimus} and StarCraft II \cite{ma2024large}, yet early work often relied on privileged APIs for simplified observations and semantic action spaces \cite{wang2023voyager}, limiting applicability to closed-source commercial games that expose only raw pixels. Concurrently, the integration of LLMs \cite{achiam2023gpt} into multi-agent systems has established a new architectural paradigm in which agents actively reason, communicate, and interact with external tools, environments, and users through structured interfaces \cite{he2025llm}. This paradigm elevates LLMs from passive text generators to dynamic participants capable of orchestrating complex tasks \cite{zhang2024large}. Notably, the rise of multimodal capabilities has spurred the development of GUI-based agents \cite{zhang2025api}, e.g., OpenAI Operator, UFO \cite{zhang2024ufo}, and CogAgent \cite{hong2024cogagent}, which integrate visual understanding with text-based reasoning to enable general and intuitive control. In practice, modern LLM-based agents function as autonomous or semi-autonomous entities that combine natural language understanding with tool-augmented reasoning \cite{zhao2025llm}, with GUIs further extending their interactive capacity \cite{tang2025survey}. In simulated or game environments, GUI perception modules can translate visual elements (e.g., buttons, maps, menus) into semantic representations for LLM reasoning \cite{hu2025agents}, enabling operation across diverse digital environments without task-specific policy training \cite{tang2025dsgbench}. Frameworks such as AutoGen \cite{wu2024autogen}, MetaAgent \cite{li2023metaagents}, and Generative Agents \cite{park2023generative} formalize these patterns, offering abstractions for agent roles, tool integration, message routing, and environment interfacing. By unifying natural language communication with API-driven and GUI-mediated interaction, LLM-based multi-agent systems are advancing from conceptual models to deployable platforms for open-ended, human-aligned reasoning and decision-making.

\textbf{API-Free LLM-based Agents.} While early LLM-based agents frequently depended on privileged access to internal game states via APIs, this API-centric paradigm faces critical constraints \cite{wang2023voyager}. Its reliance on engineered interfaces inherently limits generalization, particularly in closed-source commercial games and proprietary software where internal APIs are inaccessible or undocumented \cite{tan2025cradle}, and distances the agent from the rich visual context of genuine human-computer interaction. In pursuit of ultimate generality, a new paradigm has emerged: API-free LLM-based agents that operate purely from raw pixel input, interacting through universal human-style interfaces without relying on internal state APIs \cite{tan2025cradle, du2025rethinking}. Exemplified by systems like CRADLE \cite{tan2025cradle} and the Bottom-Up Agent \cite{du2025rethinking}, these agents learn complex skills from scratch via an autonomous trial-and-error loop. This approach has demonstrated tangible progress in open-ended environments, such as multimodal desktop assistants (e.g., UFO \cite{zhang2024ufo}, CogAgent \cite{hong2024cogagent}) and game-playing agents, achieving meaningful in-game progress without game-specific adaptations. However, this promising paradigm introduces a critical challenge: \textit{experience siloing}. Agents often retrieve memory in a localized and myopic way, relying on the most visually similar past state, and consequently fail to generalize across functionally similar but visually different contexts. This forces them to fall back on inefficient trial-and-error relearning, which hinders systematic knowledge accumulation and long-term strategy formation, underscoring a key obstacle to achieving truly general and flexible AI systems.

\section{Methodology}
\label{sec:methodology}
\begin{figure}[t]
\centering
\includegraphics[width=\linewidth]{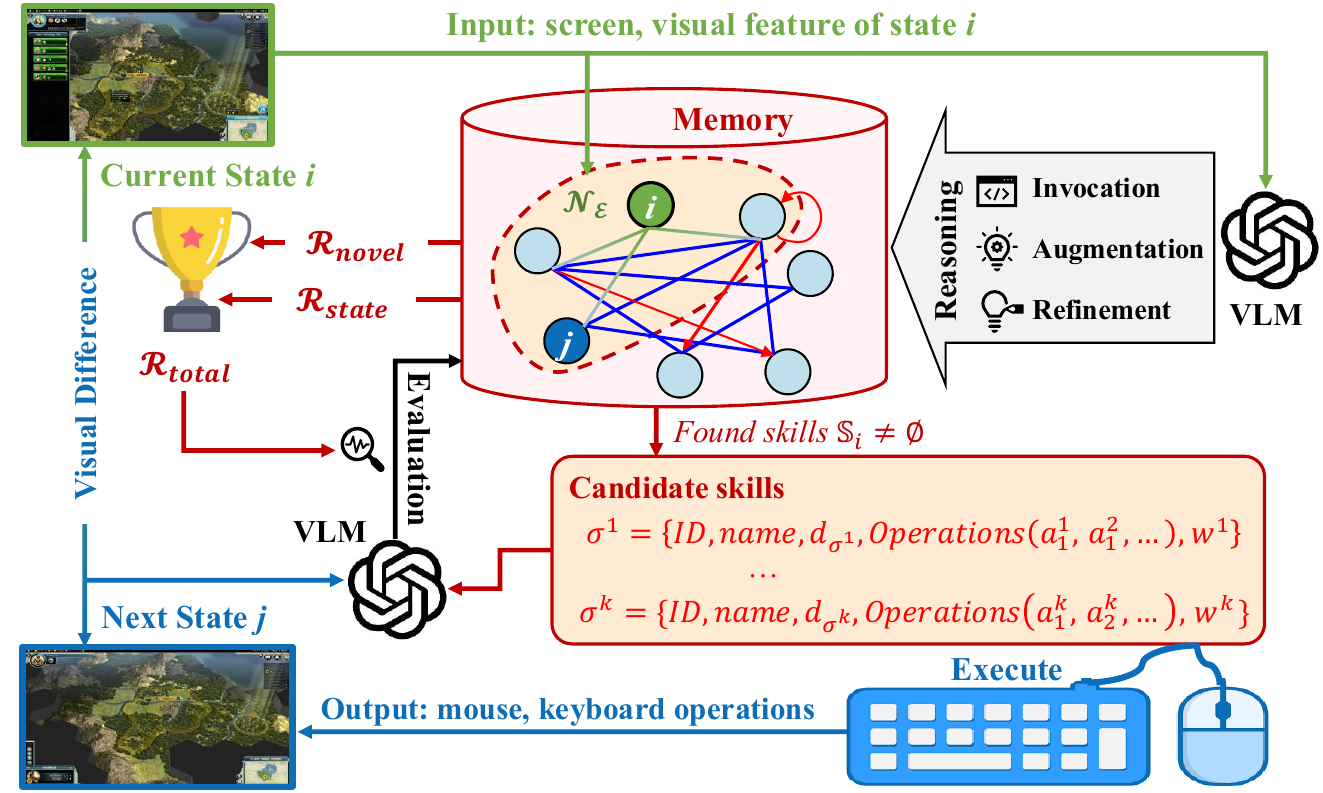}
\caption{Overview of the proposed SAG-Agent. For a given state $i$, the agent queries its memory to find a neighborhood of experience ($\mathcal{N}_\mathcal{E}$), where state nodes are connected by similarity edges (blue line) or skill edges (red arrow). If candidate skills $\mathbb{S}_{i}$ are retrieved, they are executed to transition the environment to state $j$. In addition, a VLM is central to this agent's reasoning, performing skill invocation, refinement, augmentation, and evaluation.}
\label{fig:method}
\vspace{-1em}
\end{figure}

As shown in Fig. \ref{fig:method}, the proposed \textbf{SAG-Agent} is composed of an Environment IO Interface for environment interaction, a Memory System to store and structure experience, and a VLM-based Reasoning Module that directs the agent's behavior. Following Bottom-Up Agent \cite{du2025rethinking}, we model the environment as a Partially Observable Markov Decision Process \cite{spaan2012partially} defined by $(\mathcal{X}, \mathcal{A}, \mathcal{T}, \mathcal{R})$, where $\mathcal{X}$ is the visual observation space, $\mathcal{A}$ is the set of atomic actions, $\mathcal{T}$ represents the unknown transition dynamics, and $\mathcal{R}$ is the implicit reward signal. The agent is built around a VLM $f_{*}$ that, when conditioned on a prompt $\mathbb{P}$ and current context $c_i$, produces function-specific outputs: $f_{*}(\mathbb{P}, c_i) \rightarrow y_i$. A skill $\sigma$ is defined as a sequence of atomic actions $\sigma = (a_1, …, a_k)$, each paired with a semantic descriptor $d_\sigma$, a natural language summary of its intent, generated by the VLM. The skill library $\mathbb{S} = \{\sigma_1, …, \sigma_n\}$ evolves over time through discovery, refinement, and composition. 

\subsection{The Environment IO Interface}
\label{sec:io}
To operate in API-free environments, our agent requires a universal interface for perception and action. The Environment IO Interface, fulfills this role, serving as the sole conduit through which the agent perceives and interacts with its environment. It operates exclusively on raw visual input and produces low-level, human-like actions, without access to internal states or privileged APIs. This design ensures our agent's generalizability while posing significant challenges in visual understanding and precise action execution.

\textbf{Environment Input: Purely Visual Observations.} The agent's perception is grounded exclusively in raw visual observations. At each time step $i$, the agent receives a screenshot $X_i \in \mathcal{X}$, which constitutes its entire sensory input. From this screenshot, a feature vector $x_i$ is extracted using the image encoder of a pre-trained CLIP model \cite{radford2021learning} to represent the state. Notably, our agent forgoes the use of any Optical Character Recognition (OCR) models to extract textual content. This forces the agent to rely entirely on the VLM’s intrinsic spatial perception and visual understanding to interpret the scene, without access to structured information.

\textbf{Environment Output: Simulated Mouse and Keyboard Operations.} The agent interacts with the environment by generating low-level keyboard and mouse commands from the action space $\mathcal{A}$, mirroring human interaction. The action space includes all possible operations such as \textit{click}, \textit{drag}, \textit{scroll}, and \textit{type}. These can be combined in various ways to form combos and shortcuts. Considering that mouse operations can theoretically target any pixel on the screen, making the action space excessively large, we employ the Segment Anything Model (SAM) \cite{kirillov2023segment} to dynamically identify and segment interactable UI elements from the current observation $X_i$. The interactable UI elements are also updated in the memory. A skill $\sigma$ is typically composed of one or more actions parameterized by the objects of interaction. For example, the skill '\textit{Choose Production and Build Monument}' translates to the operations: \textit{\{'operate': 'Click', 'object\_id': 11, 'object\_name': 'Choose\_Production'\}; \{'operate': 'Click', 'object\_id': 24, 'object\_name': 'Monument'\}}. To execute such an operation, the agent first retrieves a reference image of the target object (e.g., the 'Choose\_Production' button) from memory. It then performs template matching against the current screen $X_i$ to find the object's precise location, calculating the center coordinates for the mouse click. This process grounds the VLM's symbolic action plan into concrete, executable operating system-level keyboard and mouse operations on the GUI.

\subsection{Experience-Driven Memory System}
\label{sec:memory}
The memory of SAG-Agent is designed to store and manage all useful information, enabling the agent to learn from past experiences and make informed decisions. It consists of two components: a Procedural Memory and an SAG.

\textbf{Procedural Memory.} Inspired by Bottom-Up Agent \cite{du2025rethinking}, this component maintains the agent's immediate, operational knowledge, including executable skills and cached plans. It is implemented as a set of structured tables: i) \textit{Objects Table}: Stores information about interactable UI elements identified by SAM, including their names and reference images. ii) \textit{Skill Table}: Contains the evolving skill library $\mathbb{S}$, where each skill $\sigma$ is defined by its name, semantic descriptor $d_\sigma$, a sequence of operations $\{a_1, a_2, ..., a_k\}$, and a fitness score $\phi_\sigma$ that reflects its historical effectiveness. iii) \textit{Action Clusters Table}: Groups skills that are applicable in similar states (identified by their feature vectors), facilitating efficient retrieval of contextually relevant skills. iv) \textit{Monte Carlo Tree Search (MCTS) Table}: Stores serialized search trees \cite{browne2012survey} associated with specific states, allowing the agent to cache and reuse complex decision-making plans.

\textbf{State-Action Graph (SAG).} The SAG is the cornerstone of the agent's long-term memory, designed specifically to overcome the bottleneck of inefficient exploration. It is a graph $\mathcal{G}=(\mathcal{V}, \mathcal{E})$ that transforms the agent’s memory from a passive store of episodes into a semantically connected network that facilitates strategic reasoning and generalization. The SAG is constructed from nodes $\mathcal{V}$ and two types of edges: similarity edges $\mathcal{E}^{Sim}$ and skill edges $\mathcal{E}^{\sigma}$: i) \textit{Nodes ($\mathcal{V}$)}: Each node $v_i \in \mathcal{V}$ represents a unique state, characterized by its CLIP feature vector $x_i$. A new observation with feature vector $x_j$ is merged into an existing node $v_i$ if their cosine similarity exceeds a threshold, i.e., $\cos(x_i, x_j) > \theta_{{merge}}$; otherwise, a new node $v_j$ is created. ii) \textit{Similarity Edges ($\mathcal{E}^{Sim}$)}: An undirected edge $e^{\mathrm{sim}}_{i,j}$ connects two functionally analogous but visually distinct states $v_i$ and $v_j$ if $\theta_{{merge}} > \cos(x_i, x_j) > \theta_{{simi}}$. Each edge has a weight $w^{\mathrm{sim}}_{i,j} \in [0,1]$ equal to the cosine similarity between $x_i$ and $x_j$. iii) \textit{Skill Edges ($\mathcal{E}^{\sigma}$)}: A directed edge $e^{\sigma}_{i,j}$ from state $v_i$ to $v_j$ represents the successful execution of a skill $\sigma$. Each skill edge has a weight $w^{\sigma}_{i,j}$ quantifying the utility of that skill $\sigma$ for the specific state transition. Crucially, some actions produce a large immediate visual change but lead to a dead end, while a crucial setup move might have minimal impact but be key to a long-term strategy. To capture this, we define the skill edge weight $w^{\sigma}_{i,j}$ as a combination of immediate visual change and historical effectiveness:
\begin{equation}
w^{\sigma}_{i,j} = f_{sigmoid}\left(\alpha \Delta_{i,j} + (1-\alpha) \frac{\phi_\sigma}{\phi_\sigma+C_0}\right),
\label{eq:skillweight}
\end{equation}
where $f_{sigmoid}(\cdot)$ is the Sigmoid activation function, the $\Delta_{i,j}$ is the visual change ratio between states $v_i$ and $v_j$, the $\phi_\sigma$ is the historical fitness of skill $\sigma$, the $C_0$ is a constant that controls the sensitivity of the fitness term, and the $\alpha$ is a weighting factor that balances the importance of immediate change versus long-term strategic value. We quantify visual change as the proportion of changed pixels between consecutive grayscale screenshots. Pixels with an absolute difference exceeding 30 (0–255 scale) are counted, and the ratio $\Delta_{i,j}$ of such pixels to total pixels is computed. This design allows the agent to develop sophisticated, long-horizon plans by valuing both types of actions.

\subsection{Reasoning and Decision-Making Module}
\label{sec:reason}
The reasoning module is the decision-making engine of SAG-Agent, responsible for a continuous cycle of skill invocation, augmentation, refinement, and evaluation. It orchestrates a hierarchical decision-making process that leverages the structured knowledge in both the SAG and Procedural Memory to intelligently balance exploiting known strategies with exploring new possibilities. This entire process is guided by a novel, graph-based hybrid reward mechanism.

\textbf{Skill Invocation.} The SAG-Agent employs a hierarchical, two-stage process for skill invocation that prioritizes structured, experience-based knowledge from the SAG before falling back to a more general VLM-guided search, ensuring both efficiency and reliability. Upon entering a new state $v_i$, the agent first identifies its \textit{Neighborhood of Experience} $\mathcal{N}_\mathcal{E}(v_i)$, defined as the set of all nodes connected to $v_i$ by similarity edges:
\begin{equation}
\label{eq:neighborhoodexperience}
\mathcal{N}_\mathcal{E}(v_i) = \{v_j|e_{i,j}^{sim}\}.   
\end{equation}
From this \textit{Neighborhood of Experience} $\mathcal{N}_\mathcal{E}(v_i)$, it gathers a set of high-quality candidate skills $\mathbb{S}_{HQ}(v_i)$ by collecting all skills $\{\sigma_k\}$ found on outgoing skill edges from every node:
\begin{equation}
\label{eq:highqualityskill}
\mathbb{S}_{HQ}(v_i) = \bigcup_{v_j \in \mathcal{N}_\mathcal{E}(v_i)} \{\sigma_k \mid \exists v_l \text{ such that } e_{j,l}^{\sigma_k}\},
\end{equation}
where $e_{j,l}^{\sigma_k}$ represents a skill edge from a neighboring node $v_j$ to another node $v_l$ that is associated with skill $\sigma_k$. A probability distribution is then constructed to select a skill for execution. The probability $P(\sigma_k|v_i)$ of selecting a specific skill $\sigma_k$ from the candidate set $\mathbb{S}_{HQ}(v_i)$ is directly proportional to its corresponding weight, which we denote as $w^{\sigma_k}$. To form a valid probability distribution, this is normalized by the sum of all weights for all high-quality candidate skills:
\begin{equation}
\label{eq:sampleskillfromkg}
P(\sigma_k | v_i) = \frac{w^{\sigma_k}}{\sum_{w^{\sigma_l} \in \mathbb{S}_{HQ}(v_i)} w^{\sigma_l}}.
\end{equation}
From the distribution $P(\sigma_k | v_i)$, the agent samples up to $M$ skills for execution attempts. If this primary strategy fails to yield any successful skills, the agent performs a VLM-guided skill invocation from its Procedural Memory. This process queries the \textit{Action Clusters Table} to retrieve a set of candidate skills relevant to the current context. Formally, this is represented as:
\begin{equation}
\mathbb{S}_{C}(v_i) = \{\sigma_k \mid \sigma_k \in \mathbb{S} \ \land \ f_{*}(\mathbb{P}^{invoke}, X_i, \mathbb{S}) = \sigma_k\},
\end{equation}
where $\mathbb{S}_{C}(v_i)$ denotes the candidate skills retrieved by the VLM based on the visual observation $X_i$. To avoid committing greedily in a stochastic and partially observable environment, the agent evaluates the candidate skills $\mathbb{S}_{C}(v_i)$ using an approach inspired by the Upper Confidence bound for Trees (UCT) algorithm \cite{couetoux2011continuous}, a core component of MCTS. For each candidate skill $\sigma_k \in \mathbb{S}_{C}(v_i)$, the agent calculates a potential utility score which balances exploitation of known, high-value skills with the exploration of less-tried options. This Upper Confidence Bound (UCB) score $\eta_{\sigma_k}$ is defined as:
\begin{equation}
\eta_{\sigma_k} = \phi(\sigma_k) + C_1 \sqrt{\frac{\ln N_i}{n_k}} - P(\sigma_k),
\label{eq:uct}
\end{equation}
where $\phi(\sigma_k)$ quantifies the fitness of skill $\sigma_k$. Initially set to 0, its value is updated based on two factors: detectable visual state transitions following execution, and the consistency of the skill’s semantic description with the outcomes observed across before-and-after frames. The second term is the exploration bonus, where the $n_k$ is the execution count for skill $\sigma_k$, the $N_i$ is the total number of selections $\mathbb{S}_{C}(v_i)$, and the $C$ is a constant controlling the exploration-exploitation trade-off. $P(\sigma_k)$ is a penalty term that dynamically reduces the utility of skills whose prerequisite objects are not currently available, promoting the selection of fully completable actions. Finally, instead of deterministically picking the skill with the highest utility, the agent converts these scores into a probability distribution using a temperature-scaled softmax function. The probability of selecting skill $\sigma_k$ is given by:
\begin{equation}
\label{eq:sampleskill}
P'(\sigma_k|v_i) = \frac{\exp(\eta_{\sigma_k} / \tau)}{\sum_{\sigma_l \in \mathbb{S}_{C}(v_i)} \exp(\eta_{\sigma_l} / \tau)}.
\end{equation}
The skill to be executed is then stochastically sampled from this distribution. The temperature $\tau$ dynamically adjusts the randomness of the selection. This method ensures a robust balance between exploiting proven strategies and exploring new possibilities. If the candidate set $\mathbb{S}_{C}(v_i)$ is empty, the agent defaults to skill augmentation in the current context.

\textbf{Skill Augmentation \& Refinement.} 
In open-ended environments, where predefined APIs and priors are absent, most atomic actions $a \in \mathcal{A}$ are inherently task-irrelevant and semantically ambiguous. Consequently, discovering useful skills $\sigma = (a_1, \ldots, a_k)$ necessitates a structured trial-and-reasoning process, in which the agent explores diverse action combinations and evaluates their outcomes to identify meaningful behaviors. To manage the intractably large action space, we adopt the strategy by leveraging SAM to identify and segment UI elements, as well as constructing skills through incremental increases in sequence length $k$. Notably, our skill augmentation approach is driven by successful atomic action creation and validation rather than exhaustive combinatorial search. We partition the atomic action set $\mathcal{A}$ into three ordered subsets, i.e., $\mathcal{A}_1$, $\mathcal{A}_2$, and $\mathcal{A}_3$, to optimize the search space and accelerate exploration. The agent begins with $\mathcal{A}_1$, which contains previously validated single-step skills, then progresses to $\mathcal{A}_2$, comprising actions on recognized UI objects that prioritize semantically meaningful interactions, and finally considers $\mathcal{A}_3$, which includes residual actions such as clicks on unrecognized or background regions. The skill construction process starts with single-step actions ($k = 1$) and incrementally expands to longer sequences ($k = 2, 3, \ldots$). Each candidate skill $\sigma_k$ is formed by appending a new atomic action $a_k$ to a validated shorter skill $\sigma_{k-1}$. Skill expansion terminates as soon as an action sequence produces any recognizable and valuable effect, e.g., GUI transitions, measurable task progress, or meaningful state changes in the SAG—at which point the skill is annotated functionally and stored in memory.

To maintain an efficient and scalable skill library, we implement a dynamic skill pruning mechanism. MCTS is employed to ensure all skills receive sufficient testing opportunities. During skill execution, the agent records trajectory data and computes UCB scores according to Eq. \ref{eq:uct}. Skills that exceed the average visitation count and consistently demonstrate the lowest UCB scores are pruned from memory. This iterative process of expansion and pruning ensures that the skill library retains only high-utility behaviors, enabling complex, adaptive strategies to emerge compositionally from primitive actions. Furthermore, to maintain a compact and non-redundant skill set, the agent periodically triggers the VLM to consolidate the skill repertoire by clustering and merging functionally equivalent skills. This process is defined as:
\begin{equation}
\label{eq:refineskill}
    \sigma' = f_{*}(\mathbb{P}^{refine}, (X_i, \sigma, \mathcal{T})).
\end{equation}
The dual process, i.e., combining population-based pruning with LLM-guided skill consolidation, ensures that $\mathbb{S}$ undergoes evolution towards a more reusable and semantically coherent state by eliminating redundancy and promoting general-purpose skills.

\begin{algorithm}[t]
\caption{Skill Evolution of SAG-Agent.}
\label{alg:SAG-Agent}
\KwIn{Environment $(\mathcal{X}, \mathcal{A}, \mathcal{T}, \mathcal{R})$, VLM $f_{*}$, Skill library $\mathbb{S} = \emptyset$, and SAG $\mathcal{G}(\mathcal{V,E}) = \emptyset$.}
\For{Each Agent}{
    \While{Episode not terminated}{
        Observe screen $X_{i}$ and extract feature $x_{i}$\;
        Generate node $v_{i}$ and similarity edges $e^{sim}$\;
        Identify \textit{Neighborhood of Experience} by Eq. \ref{eq:neighborhoodexperience}\;
        Gather \textit{High-Quality Candidate Skills} by Eq. \ref{eq:highqualityskill}\;
        \eIf{$\mathbb{S}_{HQ}(v_i) \neq \emptyset$}{
            \For{$attempt = 1$ to $max\_attempts$}{
                Sample high-quality skill $\sigma_k$ by Eq. \ref{eq:sampleskillfromkg}\;
                Compute $\mathcal{R}_{total}$ by Eq. \ref{eq:reward}\;
                Generate skill edges $w^{\sigma_k}_{i,j}$ by Eq. \ref{eq:skillweight}\;
                \If{$\mathcal{R}_{total}(v_i,v_j) > threshold $}{
                    \textbf{Break}\;
                }
            }
        }{
            Sample skill $\sigma_l$ from $\mathbb{S}_C(v_i)$ by Eq. \ref{eq:sampleskill}\;
            \eIf{$\mathbb{S}_C(v_i) = \emptyset$}{
                \For{$k = 1$ to $k_{max}$}{
                    Gen. $a \leftarrow f_{*}(\mathbb{P}^{augment},X_i,\mathbb{O})$\;
                    \If{$a$ is $recognizable$}{
                        Update Skill library $\mathbb{S}$\;
                        \textbf{break}\;
                    }
                }
            }{
                Select skill cluster by $f_{*}(\mathbb{P}^{invoke},X,\mathbb{S})$\;
                Eval. the skill cluster with MCTS\;
                Execute the best skill $\sigma$ from MCTS\;
            }
            Observe $\mathcal{T}$ \& compute $\mathcal{R}_{total}$ by Eq. \ref{eq:reward}\;
            \If{$\mathcal{R}_{total}(v_i,v_j) < threshold$}{
                Remove $\sigma$ from $\mathbb{S}$\;
            }
            \If{$skill\_count < threshold$}{
                Refine skill: $\sigma' \leftarrow f_{*}(\mathbb{P}^{refine}, X, \sigma, \mathcal{T})$\;
                Replace $\sigma$ with $\sigma'$ if improvement\;
            }
        }
    }
}
\end{algorithm}

\textbf{Skill Evaluation.} 
For any skill $\sigma = (a_1, \ldots, a_k)$ that transitions the agent from state $v_i$ to $v_j$, the agent receives a resulting trajectory $\mathcal{T}$ from which it derives a behavioral signal. Since there is no external reward, the quality of the skill is implicitly evaluated via $\mathcal{R}_{total}$:
\begin{equation}
\label{eq:reward}
    \mathcal{R}_{total} = \mathcal{R}_{progress} + \mathcal{R}_{semantic} + \mathcal{R}_{state}+\mathcal{R}_{novel},
\end{equation}
where $\mathcal{R}_{progress}$ evaluates the progress of the game via VLM to encourage the skill to lead to specific progress in the game, $\mathcal{R}_{semantic}$ measures the consistency via VLM between the predicted high-level effects of the skill and the actual results observed in the environment, $\mathcal{R}_{state}$ encourages the agent to execute skills that lead to states with greater future potential, $\mathcal{R}_{novel}$ serves as an intrinsic curiosity drive, directly rewarding the agent to explore and expand its knowledge. 

To compute $\mathcal{R}_{state}$, we define a static skill potential as the sum of the weights of all outgoing skill edges. Then, the state-value reward for a transition from $v_i$ to $v_j$ is defined as the improvement in the agent's estimated long-term potential:
\begin{equation}
\label{eq:statereward}
\mathcal{R}_{state} = \sum_{e \in \mathcal{E}^{\sigma}(v_j)} w^\sigma(e) - \sum_{e \in \mathcal{E}^{\sigma}(v_i)} w^\sigma(e).
\end{equation}
A positive reward encourages movements to states from which higher future returns are expected. On the other hand, the novelty reward is a binary intrinsic incentive based on the prior discovery of a state. Let $\mathcal{V}_{G}$ be the set of all state vertices in the KG. The reward is defined as:
\begin{equation}
\label{eq:novelreward}
\mathcal{R}_{novel}(v_j) =
\begin{cases}
1.000, & \text{if } v_j \notin \mathcal{V}_{G} \quad \text{(new state)}, \\
0.015, & \text{if } v_j \in \mathcal{V}_{G} \quad \text{(known state)}.
\end{cases}
\end{equation}
This mechanism ensures a strong push to expand the boundaries of the known graph while providing a small, constant reward for re-visiting known states, preventing the agent from becoming completely stagnant in already-explored areas.

\subsection{Skill Evolution of SAG-Agent} 
As shown in Algorithm \ref{alg:SAG-Agent}, a core innovation of our framework is the derivation of the $\mathcal{N}_\mathcal{E}(v_i)$, which enables the agent to leverage past experience by aggregating a set of high-quality candidate skills $\mathbb{S}{HQ}(v_i)$ from the SAG, specifically, skills proven effective in analogous states. The agent then executes these skills, with the resulting reward feedback updating the corresponding skill edges in the graph. The loop terminates early upon achieving a sufficiently high reward, ensuring efficient skill reuse. When no such skills are available, the agent enters a trial-and-reasoning phase guided by a UCT-style selector within MCTS. This selector uses a fixed exploration constant $C=0.5$ and a base exploration utility threshold of $0.1$, while dynamically scheduling the temperature $\tau$ via a decay function based on the selection count of the relevant skill cluster to shift gradually from exploration to exploitation. Beyond standard UCT, the selector incorporates specialized mechanisms: a pre-selection filter excludes actions requiring unavailable or suspended objects; an action completeness penalty reduces fitness scores based on missing prerequisites; and an exploration dominance trigger activates pure exploration when the probability for exploratory actions exceeds $0.9$. These thresholds are coupled with numerical stability measures, e.g., max normalization and probability clipping in the softmax function, to ensure robust action selection in complex, partially observable environments. This integrated approach allows the SAG-Agent to dynamically balance the exploitation of proven strategies with the exploration of novel solutions, while maintaining an evolving knowledge structure that captures both state relationships and skill effectiveness.
\begin{figure}[t]
\centering
\includegraphics[width=\linewidth]{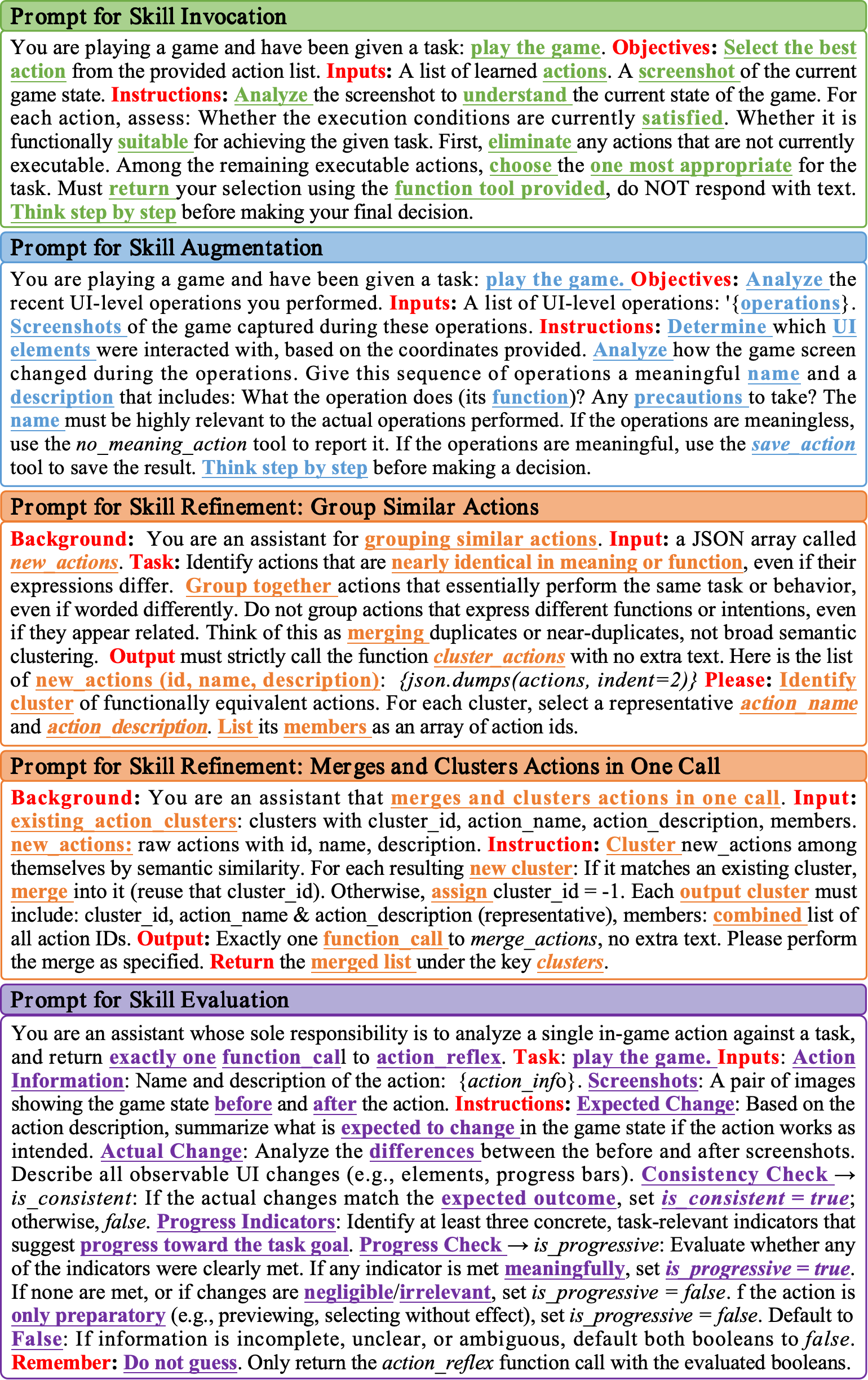}
\caption{Prompt design for VLM-based reasoning stages: skill invocation, augmentation, refinement, and evaluation.}
\label{fig:prompt}
\vspace{-1em}
\end{figure}

\begin{table*}[t]
\centering
\caption{Performance comparison across \textit{Slay the Spire} and \textit{Civilization V}. To validate performance, each agent is evaluated over three independent episodes per environment, as well as the mean and standard deviation of the final results are reported. Methods with * indicate prior-assisted setting, while those without * indicate zero-prior. NA denotes not applicable.}
\label{tab:main_results}
\resizebox{\linewidth}{!}{
\begin{tabular}{lcccccccc}
\toprule
\multirow{2}{*}{\textbf{Method}} & \multicolumn{4}{c}{\textbf{Slay the Spire}} & \multicolumn{4}{c}{\textbf{Civilization V}} \\
\cmidrule(lr){2-5} \cmidrule(lr){6-9}
 & \textbf{Floor}$\uparrow$ & \textbf{Score}$\uparrow$ & \textbf{ERR}$\uparrow$ & \textbf{Cost}$\downarrow$ &  \textbf{Turn}$\uparrow$ & \textbf{Tech.}$\uparrow$ & \textbf{ERR}$\uparrow$ & \textbf{Cost}$\downarrow$ \\
\midrule
GPT-4o* & $8.33\pm0.58$ & $48.73\pm2.43$ & $0.49\pm0.02$ & $1.01\pm0.03$ & $14.67\pm2.08$ & $1.00\pm0.00$ & $0.57\pm0.01$ & $0.91\pm0.02$ \\
Claude3.7* & $1.00\pm0.00$ & $5.00\pm0.00$ & $0.79\pm0.01$ & $1.22\pm0.04$ & $19.33\pm2.52$ & $3.33\pm0.58$ & $0.92\pm0.01$ & $1.18\pm0.09$ \\
UITARS-1.5* & $1.00\pm0.00$ & $5.00\pm0.00$ & $0.54\pm0.01$ & $0.49\pm0.24$ & $11.33\pm1.15$ & $1.33\pm0.58$ & $0.92\pm0.02$ & $0.12\pm0.01$ \\
CRADLE* & $1.00\pm0.00$ & $5.00\pm0.00$ & NA & $4.89\pm0.23$ & $0.00\pm0.00$ & $0.00\pm0.00$ & NA & $4.23\pm0.28$ \\
\midrule
GPT-4o & $1.00\pm0.00$ & $5.00\pm0.00$ & $0.72\pm0.03$ & $1.27\pm0.09$ & $6.33\pm0.58$ & $0.00\pm0.00$ & $0.37\pm0.04$ & $0.76\pm0.03$  \\
Claude3.7 & $1.00\pm0.00$ & $5.00\pm0.00$ & $0.28\pm0.01$ & $1.08\pm0.06$ & $0.00\pm0.00$ & $0.00\pm0.00$ & $0.15\pm0.3$ & $1.22\pm0.13$  \\
UITARS-1.5 & $1.00\pm0.00$ & $5.00\pm0.00$ & $0.82\pm0.01$ & \bm{$0.09\pm0.01$} & $0.00\pm0.00$ & $0.00\pm0.00$ & $0.63\pm0.01$ & \bm{$0.10\pm0.01$} \\
BottomUp & $14.33\pm1.15$ & $91.33\pm9.61$ & \bm{$0.98\pm0.01$} & $2.68\pm0.29$ & $59.67\pm8.74$ & $8.66\pm0.58$ & $0.92\pm0.01$ & $3.55\pm0.52$ \\
\rowcolor{gray!20} \textbf{SAG-Agent} & \bm{$15.67\pm0.47$} & \bm{$104.00\pm5.89$} & \bm{$0.98\pm0.01$} & $2.15\pm0.04$ & \bm{$107.33\pm6.55$} & \bm{$13.33\pm1.25 $} & \bm{$0.93\pm0.01$}  & $3.03\pm0.17 $ \\
\bottomrule
\end{tabular}}
\end{table*}

\subsection{Prompt Engineering for Agent Reasoning}
Effective reasoning in VLMs relies on meticulously designed prompts to guide complex visual understanding and decision-making processes \cite{shtedritski2023does}. In SAG-Agent, we employ a structured VLM-based reasoning cycle, i.e., skill invocation, augmentation, refinement, and evaluation. As shown in Fig. \ref{fig:prompt}, each stage is supported by a dedicated prompt that collectively enables intelligent interaction in dynamic GUI environments without reliance on pre-defined APIs. Specifically, the prompt for skill invocation selects the most suitable executable action from procedural memory when symbolic reasoning over the SAG fails. The prompt for skill augmentation complements invocation by deriving new high-level actions from low-level UI operations, thereby expanding the agent’s skill repertoire based on observed screen changes. The prompt for skill refinement consists of two sub-prompts: one groups functionally similar new actions, and the other merges them with existing skill clusters to maintain a compact and non-redundant skill library. Finally, the prompt for skill evaluation assesses the effectiveness of executed skills by comparing expected outcomes against actual screen changes, providing essential feedback for progressive and semantic rewards. Together, these prompts form a closed-loop learning architecture: invocation drives action, augmentation acquires knowledge, refinement consolidates understanding, and evaluation validates progress. 
 

\section{Experiments}
\label{sec:experiment}
\subsection{Environment and Experimental Setup}
\label{sec:setup}
\textbf{Environment.} We conduct experiments in two complex, open-ended games established by the Bottom-up Agent \cite{du2025rethinking}: \textit{Slay the Spire} (Base difficulty: Ascension 0, Character: Ironclad), a roguelike deck-builder, and \textit{Civilization V} (Civilization: Russia, Map: Earth, Size: Standard, Difficulty: Prince, Pace: Standard), a hallmark 4X strategy game. These environments serve as ideal testbeds as they demand both precise, low-level execution and sophisticated, long-horizon planning. Their turn-based nature is also well-suited to the current latency characteristics of LLM-based reasoning. An episode is terminated if the agent fails a task-specific condition (character defeat in \textit{Slay the Spire} or nation defeat in \textit{Civilization V}) or if it reaches a maximum limit of 500 steps.

\textbf{Evaluation Metrics.} We assess agent performance using a suite of four metrics designed to capture both behavioral competence and computational efficiency: i) \textit{Progression}: measured by floors cleared (\textbf{Floor}) in \textit{Slay the Spire} and turns survived (\textbf{Turn}) in \textit{Civilization V}; ii) \textit{Strategic Development}: quantified by \textit{In-Game Scores} (\textbf{Score}) in \textit{Slay the Spire} and number of technologies unlocked (\textbf{Tech.}) in \textit{Civilization V}; iii) \textit{Execution-Responsive Rate} (\textbf{ERR}): the percentage of predicted actions that lead to valid state transitions; and iv) \textit{Token Costs} (\textbf{Cost}) ($\$$) are average LLM tokens consumed per 100 steps, converted to USD for fair comparison across methods with varying episode lengths.
 
\textbf{Baselines.} We evaluate SAG-Agent against a comprehensive suite of representative baselines. This includes leading proprietary models (GPT-4o, Claude 3.7), a strong open-source GUI agent (UI-TARS-1.5 \cite{qin2025ui}), and the two most relevant API-free agents (CRADLE \cite{tan2025cradle} and the Bottom-up Agent \cite{du2025rethinking}). This selection is motivated by the limitations of many existing frameworks, which are often incompatible with the API-free setting or require task-specific priors (e.g., predefined subgoals). For the CRADLE baseline, which natively depends on a library of atomic skills, we adapted its core prompts for our tasks. To ensure a fair comparison, we converted the skills learned by SAG-Agent into a compatible format that could be utilized by the CRADLE framework, thereby providing it with the necessary prerequisite capabilities.

\textbf{Implementation Details.} We use GPT-4o for semantic reasoning and a pre-trained CLIP ViT-B/32 for visual encoding. All hyperparameters remain consistent across all experiments to ensure fair and reproducible comparisons. For the SAG, the state merging threshold $\theta_{\text{merge}}$ is set to 0.95, and the similarity edge threshold $\theta_{\text{simi}}$ to 0.88. For skill edge weight computation (Eq. \ref{eq:skillweight}), we use a balancing factor $\alpha = 0.7$ and fitness sensitivity constant $C_0 = 5.0$. The UCT-based skill selection (Eq. \ref{eq:uct}) uses an exploration constant $C_1 = 5.0$. Critically, the exact same agent implementation, including the identical hyperparameter configuration and prompt design, is deployed in both \textit{Slay the Spire} and \textit{Civilization V}. This highlights the domain-general nature of our framework, which operates effectively without environment-specific tuning.

\subsection{Comparative Results}
As shown in Table \ref{tab:main_results}, the SAG-Agent demonstrates statistically significant and consistent superiority over all baseline methods across two distinct open-ended game environments, including its zero-shot transfer to \textit{Slay the Spire} under identical architecture, prompts, and hyperparameters as used in \textit{Civilization V}, highlighting the framework's inherent adaptability and robustness without any domain-specific adjustments. Our agent achieves the highest progression metrics and maximum in-game scores while maintaining exceptional execution reliability, with performance consistency across different game genres further underscoring the robustness of our approach. The contrast with CRADLE is particularly illuminating: while CRADLE relies on extensive prior knowledge encoded in hand-designed prompts and atomic skills, it achieves only minimal progression in both games. SAG-Agent's substantial advantage without any game-specific prior knowledge validates the effectiveness of our zero-prior learning paradigm. Furthermore, our method demonstrates remarkable consistency, achieving top performance across all metrics in both games, unlike baselines that exhibit volatile performance between environments. Notably, SAG-Agent accomplishes this superior performance with reduced computational overhead, evidenced by lower token costs per 100 steps compared to the BottomUp Agent. These results confirm that our agent not only inherits BottomUp's execution reliability but significantly enhances strategic decision-making through our skill-aligned knowledge graph, ultimately affirming SAG-Agent's effectiveness in mastering complex, open-ended environments through autonomous learning and adaptation.

\begin{table}[t]
\centering
\caption{Ablation study of core components of SAG-Agent in \textit{Civilization V}, with each run constrained to 100 steps for clarity. The “w/o Simi.” means that each state forms an isolated node, with no merging or similarity-based connections permitted. $\mathcal{R}_{1}$ and $\mathcal{R}_{2}$ denote $\mathcal{R}_{progress}$ and $\mathcal{R}_{semantic}$, respectively. $\mathcal{R}_{3}$ and $\mathcal{R}_{4}$ denote $\mathcal{R}_{state}$ and $\mathcal{R}_{novel}$, respectively.}
\label{tab:ablation2}
\resizebox{\columnwidth}{!}{
\begin{tabular}{lcccccccc}
\toprule
\multirow{2}{*}{\textbf{Setting}} & \textbf{Library} & \multicolumn{3}{c}{\textbf{SAG Information}} & \multicolumn{4}{c}{\textbf{Civilization V}} \\
\cmidrule(lr){3-5} \cmidrule(lr){6-9}
 & \textbf{Size} & \textbf{Node} & \textbf{Skill} & \textbf{Simi.} &  \textbf{Turn}$\uparrow$ & \textbf{Tech.}$\uparrow$ & \textbf{ERR}$\uparrow$ & \textbf{Cost}$\downarrow$ \\
\midrule
\rowcolor{gray!20} \textbf{Full Model} & 76 & 26 & 37 & 226 & \textbf{65} & \textbf{10} & \textbf{0.89} & 3.0  \\
w/o Simi. & 30 & 672 & 62 & 0 & 15 & 1 & 0.64 & 3.5 \\
w/o $\mathcal{R}_{4}$ & 48 & 11 & 19 & 74 & 23 & 2 & 0.81 & 3.1 \\
w/o $\mathcal{R}_{3}$ & 57 & 20 & 33 & 110 & 39 & 4 & 0.87 & 3.2  \\
w/o $\mathcal{R}_{3}$ \& $\mathcal{R}_{4}$ & 64 & 18 & 39 & 96   & 35 & 3 & 0.79 & 3.3  \\
w/o $\mathcal{R}_{1}$ \& $\mathcal{R}_{2}$ & 33 & 9 & 14 & 42   & 14  & 1 & 0.59  & \textbf{2.1}  \\
\bottomrule
\end{tabular}}
\end{table}

\begin{table}[t]
\centering
\caption{Sensitivity analysis of key hyperparameters in \textit{Slay the Spire}, including the number of skills sampled per execution attempt $M$, the coefficient balancing state-value reward $\mathcal{R}_{state}$ ($\mathcal{R}_{3}$) and novelty reward $\mathcal{R}_{novel}$ ($\mathcal{R}_{3}$), and the cosine-similarity thresholds for node merging and similarity edge creation ($\theta_{merge}$, $\theta_{simi}$). \textbf{Time} is the average time in seconds required to run one step under each configuration.}
\label{tab:parameter}
\resizebox{\columnwidth}{!}{
\begin{tabular}{ccccccc}
\toprule
\multirow{2}{*}{\textbf{Configuration}} & \multicolumn{3}{c}{\textbf{SAG Information}} & \multicolumn{3}{c}{\textbf{Slay the Spire}} \\
\cmidrule(lr){2-4} \cmidrule(lr){5-7}
 & \textbf{Node} & \textbf{Skill} & \textbf{Simi.} &  \textbf{Floor}$\uparrow$ & \textbf{Score}$\uparrow$ & \textbf{Time}$\downarrow$ \\
\midrule
\rowcolor{gray!20} $M=5$   & 29   & 54   & 354   & \textbf{16}   & \textbf{112}   & \textbf{158.34s}  \\
$M=3$   & 25   & 38   & 124   & \textbf{16}   & 98   & 244.63s  \\
$M=10$   & 22   & 34   & 116   & 11   & 63   & 192.41s  \\
\midrule
\rowcolor{gray!20} $\mathcal{R}_{3}:\mathcal{R}_{4}=1:1$   & 29   & 54   & 354   & \textbf{16}   & \textbf{112}   & 158.34s  \\
$\mathcal{R}_{3}:\mathcal{R}_{4}=1:2$   & 33   & 32   & 220   & \textbf{16}   & 102   & 153.26s  \\
$\mathcal{R}_{3}:\mathcal{R}_{4}=2:1$  & 20   & 26   & 114   & 13   & 87    & \textbf{145.92s}   \\
\midrule
\rowcolor{gray!20} (0.95, 0.88)   & 29   & 54   & 354   & \textbf{16}   & \textbf{112}   & \textbf{158.34s}  \\
(0.99, 0.90)   & 155   & 190   & 14520   & 12   & 70   & 392.55s  \\
(0.85, 0.80)   & 4   & 9   & 4   & 4   & 26   & 179.40s  \\
\bottomrule
\end{tabular}}
\end{table}

\subsection{Ablation Studies}
\label{sec:ablation}
As shown in Table \ref{tab:ablation2}, the full KG‑Agent achieves the strongest performance, while removing core components leads to marked degradation. Specifically, disabling similarity-based merging (``w/o Simi.”) severely impairs long‑term progression and strategic coherence. This drop stems from a deeper architectural inconsistency: although state merging was disabled, the graph‑based novelty reward $R_{novel}$ ($R_4$) and state‑transition reward $R_{state}$ ($R_3$) remained active. The resulting mismatch caused $R_{novel}$ ($R_4$) to continuously reward the agent for registering each new state as an isolated node, promoting graph expansion over semantic exploration, while $R_{state}$ ($R_3$) operated on a fragmented graph with disconnected nodes, yielding unstable and localised value estimates. Consequently, the VLM was misled by pathological reward signals, prioritising node proliferation over meaningful progress, as reflected in the inflated node count (672 nodes vs. 26 in the full model). Similarly, ablating $\mathcal{R}{progress}$ ($R_1$) and $\mathcal{R}{semantic}$ ($R_2$) reduces the total reward available for skill evaluation, which directly weakens skill fitness $\phi(\sigma_k)$ and lowers the corresponding UCB scores during MCTS. This impairs both skill selection and pruning, resulting in notably poor outcomes across library size, graph connectivity, and in‑game progression. Together, these ablation studies confirm that our complete framework, integrating similarity‑based abstraction with a balanced multi‑reward mechanism, enables consistent state representation and stable value estimation, thereby providing reliable learning signals for effective long‑horizon planning.


\begin{table*}[t]
\centering
\caption{Evolution of skill library and SAG, as well as game performance over training rounds in \textit{Civilization V}. Each round builds on previous memory with 100 steps.}
\label{tab:ablation1}
\resizebox{0.75\linewidth}{!}{
\begin{tabular}{lcccccccccc}
\toprule
\multirow{2}{*}{\textbf{Round}} & \multicolumn{3}{c}{\textbf{Skill Library Information}} & \multicolumn{3}{c}{\textbf{SAG Information}} & \multicolumn{4}{c}{\textbf{Civilization V}} \\
\cmidrule(lr){2-4} \cmidrule(lr){5-7} \cmidrule(lr){8-11}
 & \textbf{Library} & \textbf{Augment.} & \textbf{Pruned} & \textbf{Nodes} & \textbf{Skill} & \textbf{Simi.} &  \textbf{Turn}$\uparrow$ & \textbf{Tech.}$\uparrow$ & \textbf{ERR}$\uparrow$ & \textbf{Cost}$\downarrow$ \\
\midrule
Round 0 & 76 & +77 & -1 & 26 & 37 & 226 & 65 & 10 & 0.89 & 3.0  \\
Round 1 & 106 & +33 & -3 & 50 & 84 & 782 & 98 & 12 & 0.92 & 2.8 \\
Round 2 & 111 & +7 & -2 & 53 & 90 & 856 & 113 & 13 & 0.93 & 2.7 \\
Round 3 & 114 & +4 & -1 & 55 & 90 & 838 & 115 & 13 & 0.94 & 2.7  \\
\bottomrule
\end{tabular}}
\end{table*}

\begin{figure}[t]
\small
\centering
\captionsetup[subfigure]{justification=centering}
\subfloat[Step 100]{
    \label{fig:kg-sts-100}
    \includegraphics[width=0.49\linewidth]{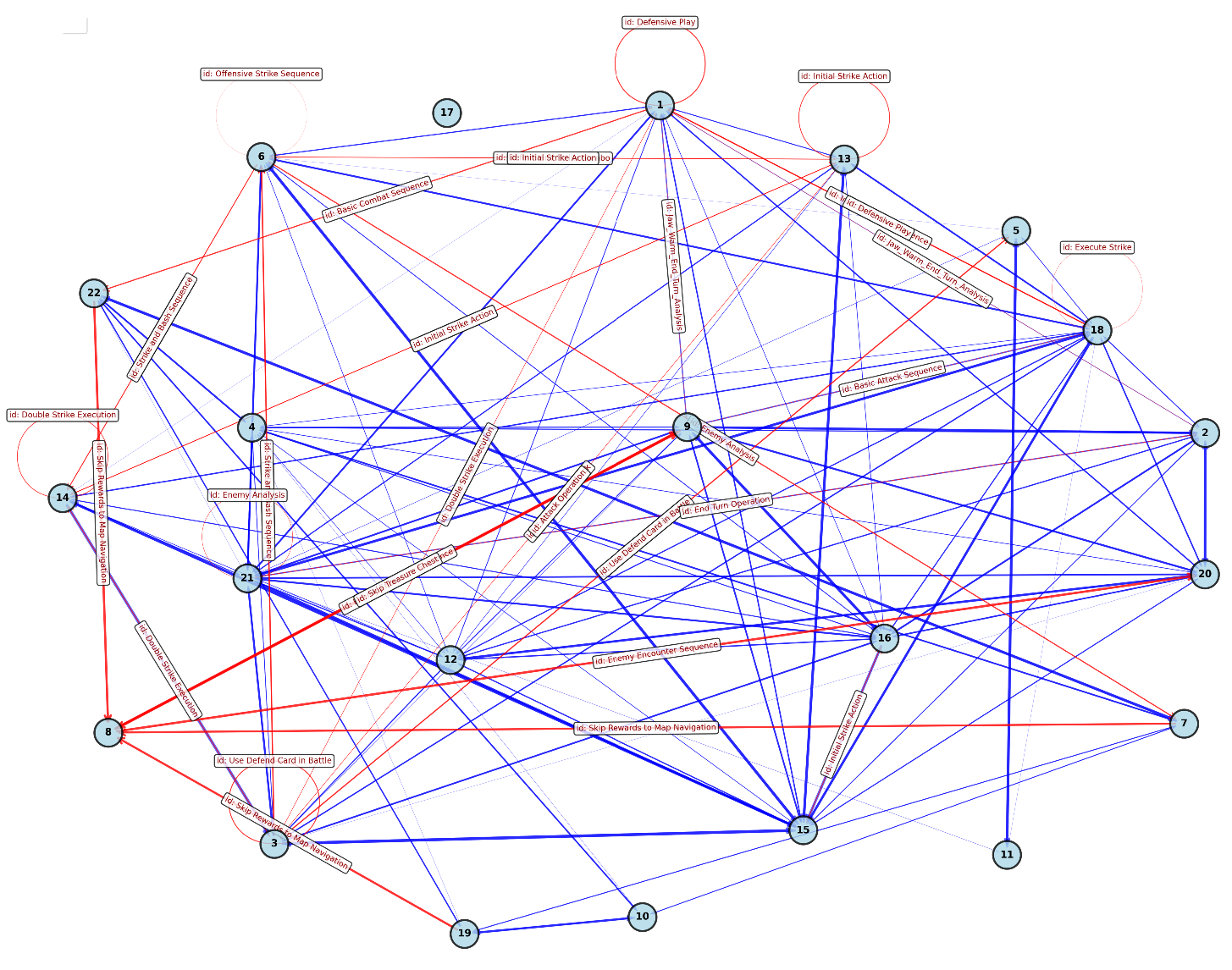}}
\subfloat[Step 200]{
    \label{fig:kg-sts-200}
    \includegraphics[width=0.49\linewidth]{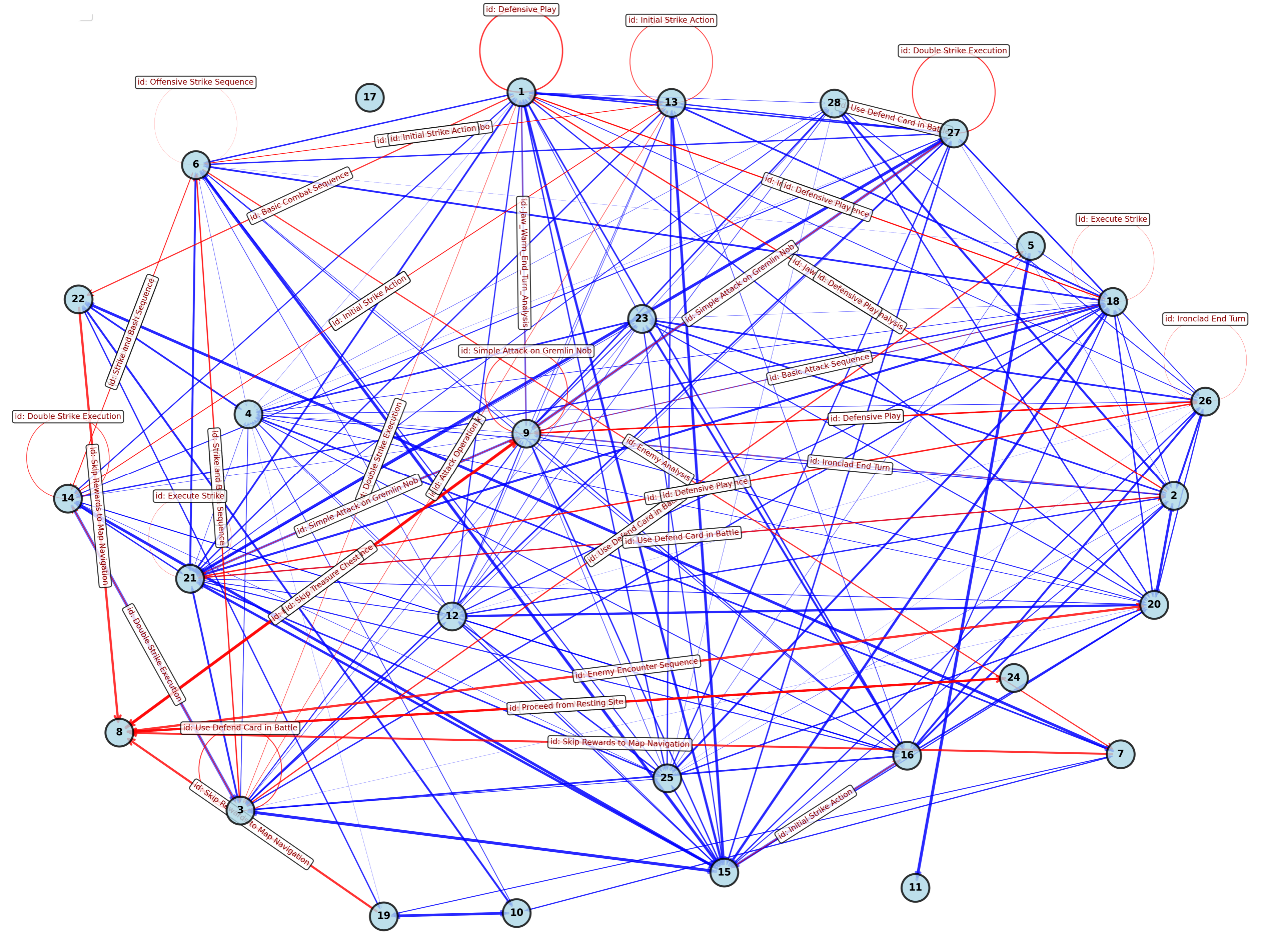}}
\caption{The SAG evolution from step 100 to 200 shows network growth in \textit{Slay the Spire}, with skill (red) and similarity (blue) connections reflecting improved structural organization and relational reasoning capabilities.}
 \label{fig:kg-sts}
 \vspace{-1em}
\end{figure}

\subsection{Model Analysis}
\label{sec:modelanalysis}
\textbf{Hyperparameter Sensitivity.} For certain hyperparameters, e.g., the criterion for increasing skill length, the fitness sensitivity and greediness weighting constants, and the temperature for skill selection, we adopt values grounded in established practices from related work \cite{du2025rethinking}. Furthermore, for the key hyperparameters directly tied to our core contributions, i.e., the number of skills sampled per execution attempt, the coefficient balancing state-value and novelty rewards, and the cosine-similarity thresholds for node merging and edge creation, we conduct a targeted sensitivity analysis in \textit{Slay the Spire}. Specifically, for each of these key parameters, we evaluated two alternative values alongside the default setting, while keeping all other configurations identical. As shown in Table \ref{tab:parameter}, our selected hyperparameter configuration, sampling a maximum of 5 skills per execution attempt ($M=5$), applying a balanced reward coefficient ($\mathcal{R}_{3}:\mathcal{R}_{4}=1:1$), and setting cosine similarity thresholds for node merging and edge creation to $\theta_{merge} = 0.95$ and $\theta_{simi} = 0.88$, stands out as the optimal setup. It achieves superior in-game performance, attaining the highest score of 112 and maximum progression of 16 floors cleared, while preserving high computational efficiency with an average runtime of 158.34 seconds per step. The choice of $M=5$ strikes an ideal trade-off between behavioral diversity and execution efficiency, outperforming both smaller and larger sampling sizes. Similarly, the 1:1 reward ratio proves more effective than imbalanced configurations, better harmonizing exploration and exploitation compared to either novelty-heavy or state-value-dominated settings. Importantly, the model exhibits robust performance across a range of configurations, with multiple hyperparameter sets yielding solid results, underscoring its general insensitivity to parameter variation, though our specific configuration delivers the best balance between task advancement and operational efficiency.
\begin{figure}[t]
\small
\centering
\captionsetup[subfigure]{justification=centering}
\subfloat[High-Reward Skills in \textit{Civilization V}]{
    \label{highreward-C5}
    \includegraphics[width=\linewidth]{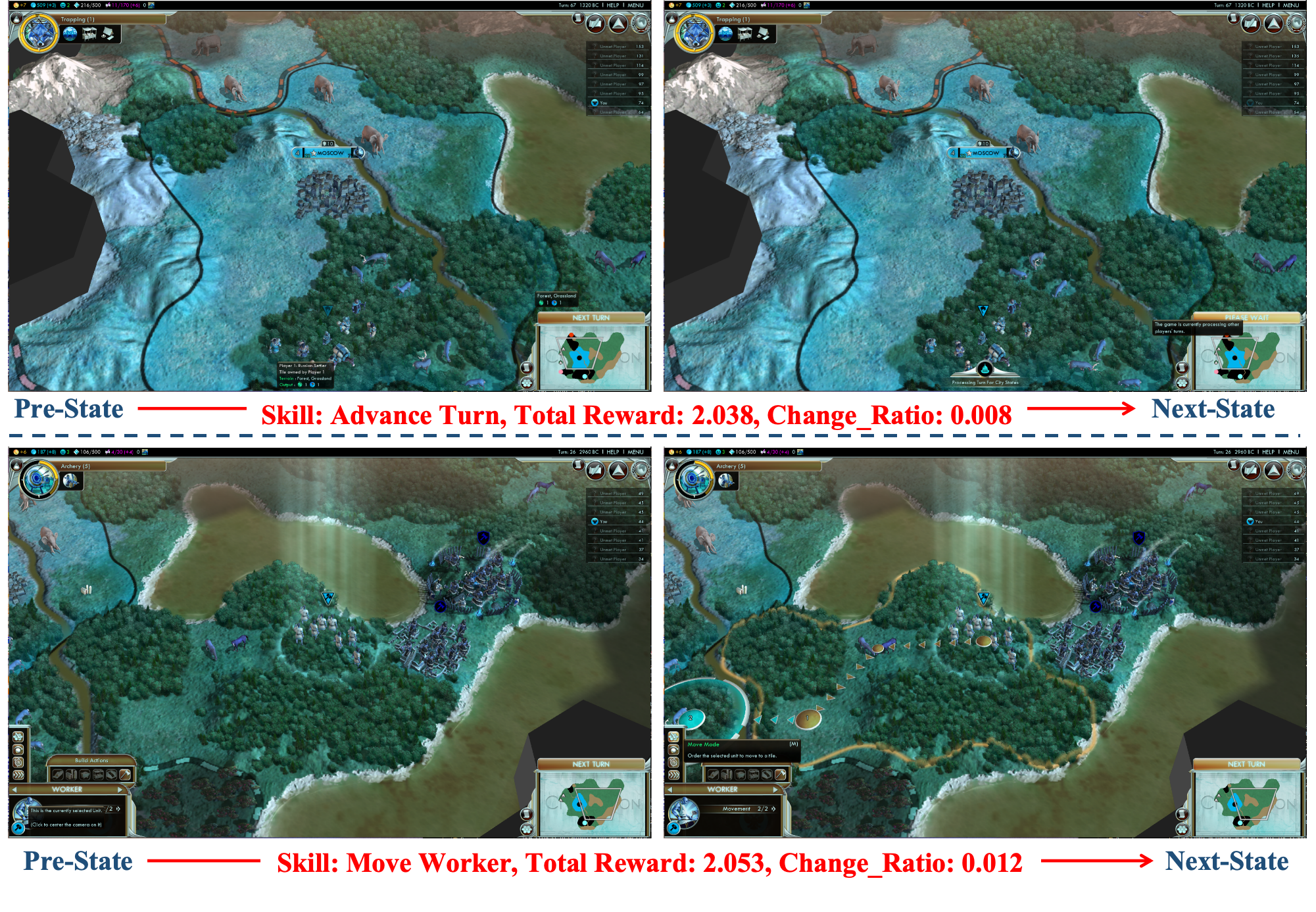}}
    
\subfloat[High-Reward Skills in \textit{Slay the Spire}]{
    \label{highreward-sts}
    \includegraphics[width=\linewidth]{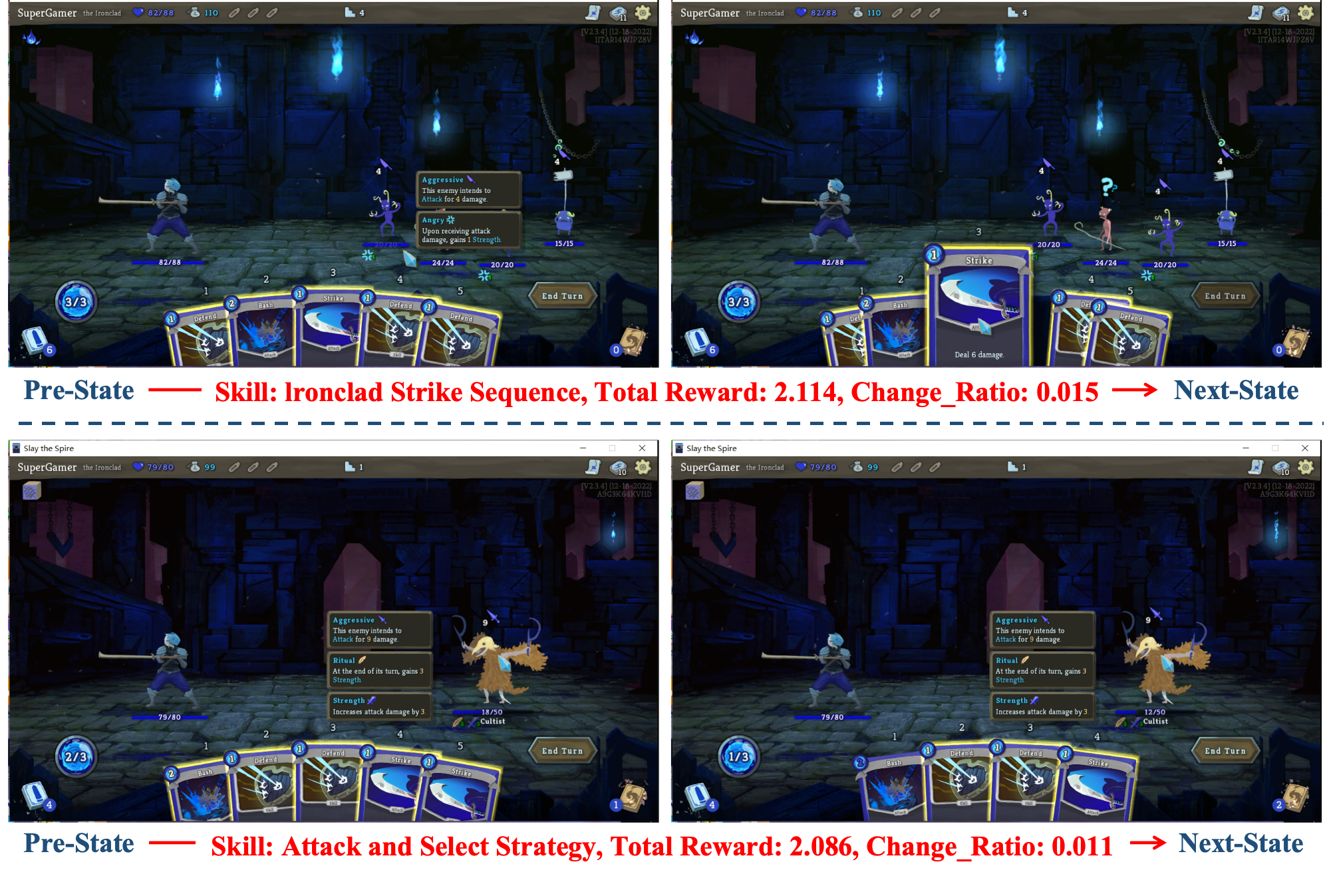}}
\caption{Our framework executes high-reward skills even with minimal visual change between states.}
 \label{fig:reward}
 \vspace{-1em}
\end{figure}

\textbf{Skill Evolution over Rounds}. The structural evolution of the skill library and SAG across training rounds is detailed in Table \ref{tab:ablation1}. In \textit{Civilization V}, the skill library grows dynamically through iterative augmentation and pruning, reflecting a continuous refinement process. Correspondingly, the SAG expands in complexity: node count increases from 26 to 55 and skill edges from 37 to 90, with both stabilizing after Round 2, indicating convergence toward a compact and functionally stable representation. Early rounds feature a marked rise in similarity edges, suggesting enhanced relational reasoning, followed by a slight decline by Round 3 as semantic consolidation and redundancy removal through functional merging take place. These structural improvements correlate with consistent gains in game progression and execution efficiency, accompanied by decreasing token costs, underscoring enhanced reasoning efficacy. By Round 3, convergence across all metrics confirms that repeated skill reuse and graph consolidation sustain agent adaptation and strategic exploration in open-ended settings. A parallel evolution is observed in \textit{Slay the Spire}, as illustrated in Fig. \ref{fig:kg-sts}. The structural evolution of the SAG from step 100 to step 200, with the evaluation terminating at step 210 after the agent’s character is defeated by a monster, exhibits notable expansion: nodes increase from 22 to 28, similarity edges from 190 to 312, and skill edges from 37 to 51. This growth, especially the accelerated formation of similarity edges, indicates improved recognition of state equivalences and environmental patterns. The steady rise in skill edges reflects effective knowledge transfer and progressive graph maturation, supported by denser connectivity and emergent cohesive clusters that facilitate robust decision-making.
\begin{figure*}[t]
\centering
\includegraphics[width=\textwidth]{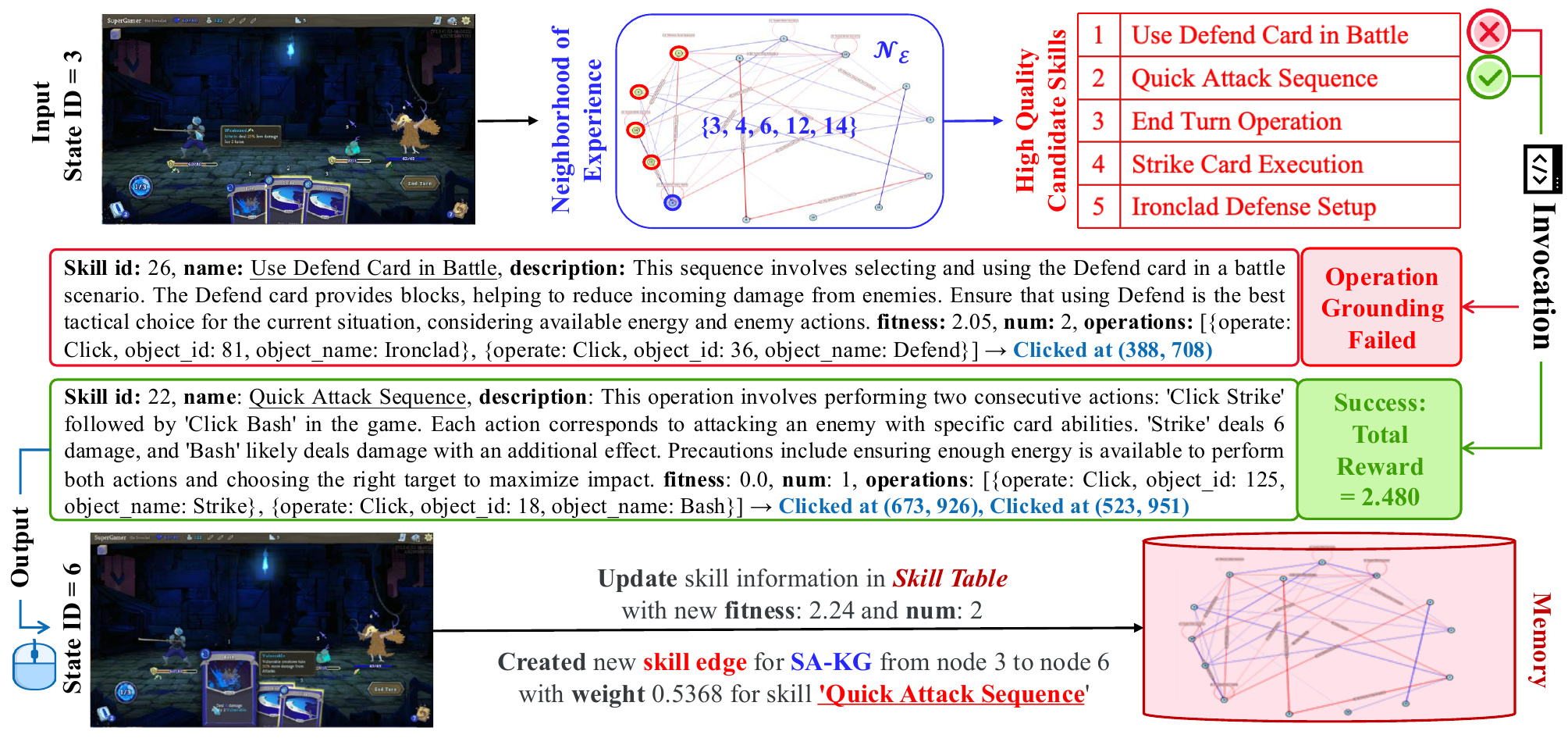}
\caption{Case study of successful skill invocation through state-conditioned sampling from the SAG.}
\label{fig:case1}
\end{figure*}

\textbf{Strategic Reasoning Beyond Perceptual Salience.} Our framework demonstrates a consistent ability to prioritize strategic value over perceptual salience, executing high-reward skills even when visual changes between consecutive states are minimal. This capability is validated in both open-ended environments, as illustrated in Fig. \ref{fig:reward}. In \textit{Civilization V} (Fig. \ref{highreward-C5}), the agent repeatedly selects skills such as ``Advance Turn” (Total Reward: 2.038, Change\_Ratio: 0.00) and ``Move Worker” (Total Reward: 2.053, Change\_Ratio: 0.012), which yield strategic progress despite negligible visual alterations, confirming that the reward mechanism can identify critical actions beyond superficial cues. Similarly, in \textit{Slay the Spire} (Fig. \ref{highreward-sts}), the agent consistently employs combat skills such as ``Ironclad Strike Sequence” (Total Reward: 2.114, Change\_Ratio: 0.015) and ``Attack and Select Strategy” (Total Reward: 2.086, Change\_Ratio: 0.011), which deliver substantial rewards while producing only subtle visual differences. These selections reflect sophisticated reasoning in card sequencing, enemy targeting, and resource management—tactical decisions that cumulatively build incremental combat advantage. The ability to recognize and execute such strategically valuable actions, even in the absence of pronounced visual feedback, underscores the framework’s capacity to model complex game dynamics and pursue long-term success through sequences of subtle, high-impact decisions.
\begin{figure*}[t]
\centering
\includegraphics[width=0.95\textwidth]{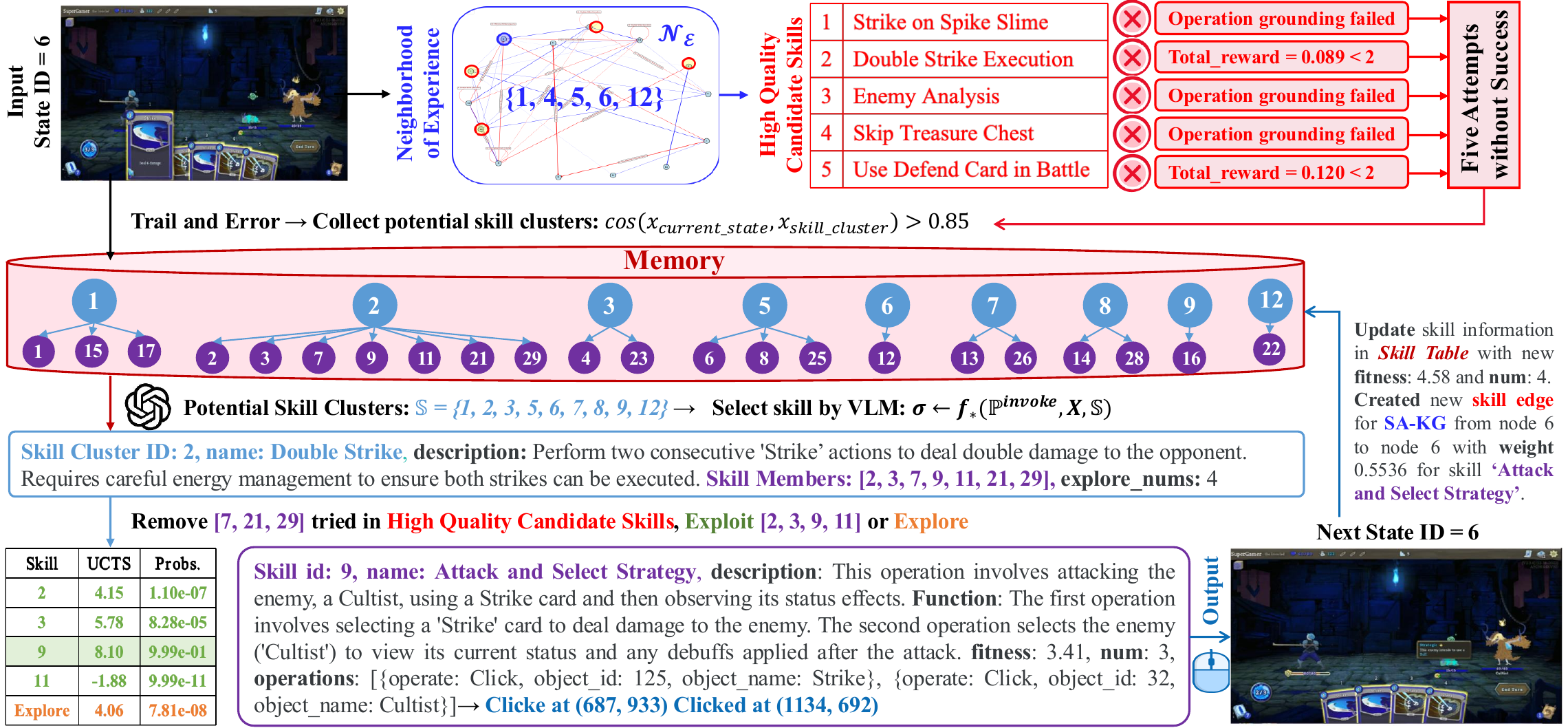}
\caption{Case study of successful skill invocation by general VLM-guided trial and error. After five unsuccessful attempts with SAG-sampled skills, SAG-Agent switches to trial-and-error mode.}
\label{fig:case2}
\end{figure*}

\subsection{Case Study} 

Fig. \ref{fig:case1} presents a case study demonstrating successful skill reuse through the SAG of SAG-Agent. When encountering a new state and corresponding node $v_3$, the proposed SAG-Agent first retrieves its \textit{Neighborhood of Experience} $\mathcal{N}_\mathcal{E}(v_3)$ and identifies high-quality candidate skills $\mathbb{S}{HQ}(v_3)$. Skills are then sampled probabilistically according to their skill edge weights in the SAG. While the initially sampled skill ``\textit{Use Defend Card in Battle}" fails due to grounding issues, the subsequently sampled skill ``\textit{Quick Attack Sequence}" executes successfully through the simulated mouse operations ``\textit{Clicked at (673, 926)}" and ``\textit{Clicked at (523, 951)}", achieving a total reward of 2.480 and transitioning to next state node $v_6$. Following successful execution, SAG-Agent updates both the fitness value and edge weight of this skill in Memory, thereby reinforcing this validated strategy for future reuse. The case study in Fig. \ref{fig:case1} illustrates how SAG-Agent addresses the key limitations of traditional LLM-based agents in API-free environments. First, by organizing pixel-level interactions into SAG, the agent overcomes inefficient exploration by generalizing across visually distinct but functionally similar states through \textit{Neighborhood of Experience}, avoiding myopic decisions. Second, the edge-weight based sampling demonstrates strategic reuse of historical knowledge rather than relying on trial-and-error. Third, the successful execution and subsequent memory update illustrate our hybrid reward mechanism: the substantial environmental reward combined with fitness updates reinforces high-value pathways while maintaining exploration flexibility. This showcases SAG-Agent's ability to decouple strategic planning from pure discovery, effectively addressing both skill acquisition and long-horizon reasoning challenges in API-free environments. 

Fig. \ref{fig:case2} demonstrates SAG-Agent's two-stage skill invocation strategy, which prioritizes structured knowledge from the SAG before resorting to VLM-guided search. Initially facing state node $v_6$, the SAG-Agent attempts SAG-based skill sampling but encounters five consecutive failures due to grounding issues or insufficient rewards. This triggers a fallback to VLM-guided skill retrieval from \textit{Procedural Memory}. The process first queries the \textit{Action Clusters Table} to obtain candidate skill clusters $\mathbb{S}_C(v_6)=\{1, 2, 3, 5, 6, 7, 8, 9, 12\}$, from which the VLM identifies the most contextually relevant cluster based on visual observation. To navigate the stochastic environment, the SAG-Agent employs a UCT-inspired strategy that balances exploitation of high-value skills $\{\sigma_2, \sigma_3, \sigma_9, \sigma_{11}\}$ with exploration of less-tried options. The resulting utility scores are converted into a probability distribution via temperature-scaled softmax, enabling stochastic sampling. The first sampled skill ``\textit{Attack and Select Strategy}" ($Probs. = 9.99e-01$) executes successfully through operations ``\textit{Clicked at (687, 933)}" and ``\textit{Clicked at (1134, 692)}", yielding a 2.117 reward. This successful execution triggers updates to the skill's fitness and edge weight in memory, demonstrating SAG-Agent's capacity for continuous knowledge refinement through both successful and failed experiences.

\section{Conclusion}
In this paper, we propose \textbf{SAG-Agent} to organize pixel-level GUI interactions into a knowledge graph of states and actions, enabling agents to overcome short-sightedness in API-free, open-ended environments. By linking functionally similar yet visually distinct states, the agent generalizes past experiences into coherent long-term strategies. Experiments in \textit{Civilization V} and \textit{Slay the Spire} demonstrate that SAG-Agent achieves significant gains in exploration efficiency and strategic depth over state-of-the-art baselines. These results highlight the potential of structuring experience into a graph-based memory, advancing API-free agents toward scalable and general-purpose autonomy. 

\textbf{Limitations and Future Work.} A key limitation of SAG-Agent is that its knowledge remains tied to specific environments, lacking the full abstraction and transferability that characterize human reasoning. While the framework structures experience into a graph-based memory, its ability to generalize knowledge across unseen tasks and diverse contexts is still constrained. Future work will focus on advancing this direction: enabling higher-level knowledge induction and reasoning that can span heterogeneous environments and ultimately testing such capabilities in more complex, real-world scenarios. This progression would move API-free agents closer to human-like adaptability and robust general intelligence.

\bibliographystyle{IEEEtran}
\begin{IEEEbiography}[{\includegraphics[width=1in,height=1.25in,clip,keepaspectratio]{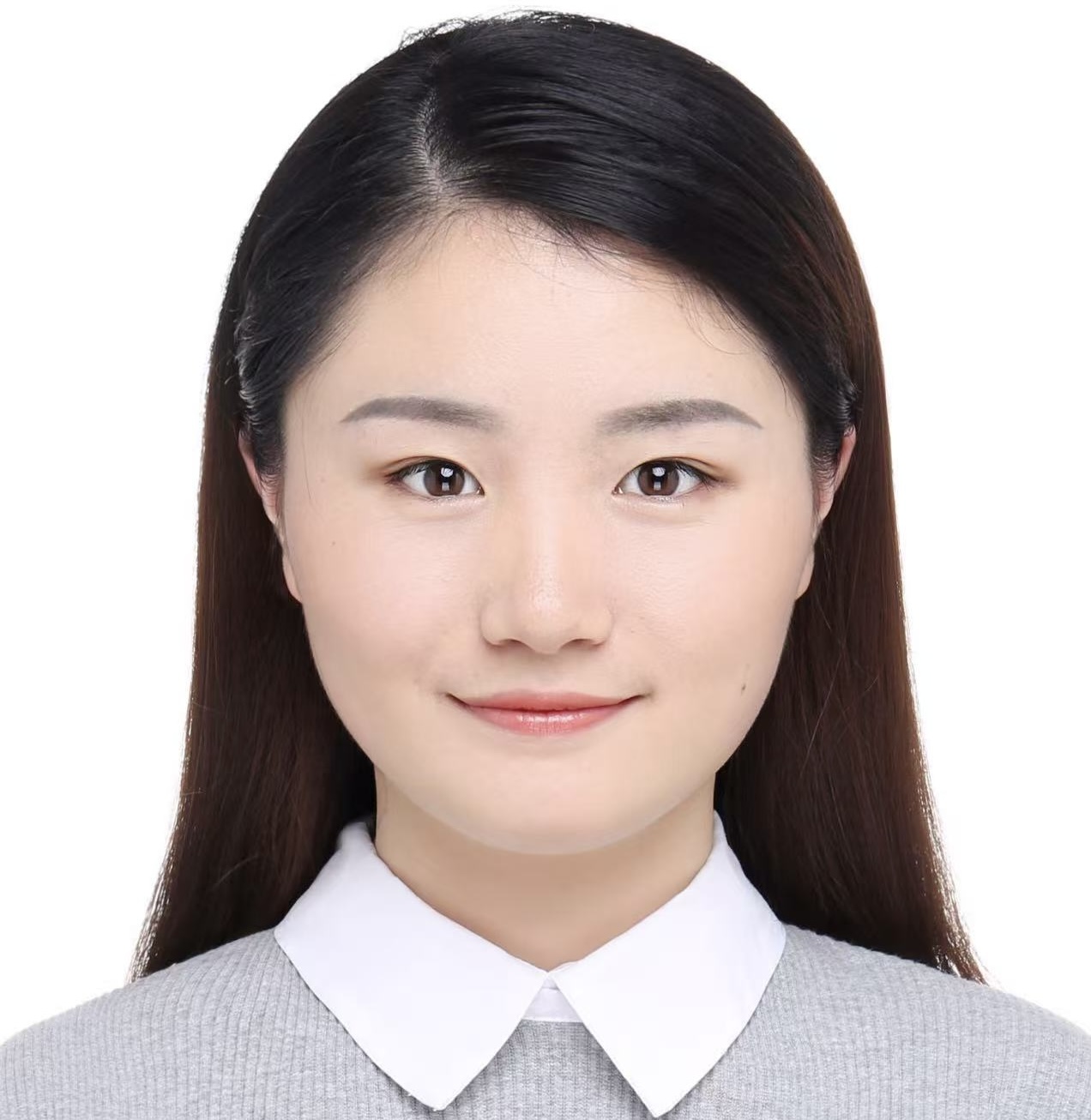}}]{Chenwei Tang} received her bachelor's and Ph.D. degrees from the College of Computer Science, Sichuan University, Chengdu, China, in 2016 and 2020. She is currently an Associate Professor with the Data Intelligence and Computing Art Laboratory, College of Computer Science, Sichuan University. Her current research interests include Agentic AI, zero-shot learning, and multi-modal learning.
\end{IEEEbiography}
\vspace{-33pt}
\begin{IEEEbiography}[{\includegraphics[width=1in,height=1.25in,clip,keepaspectratio]{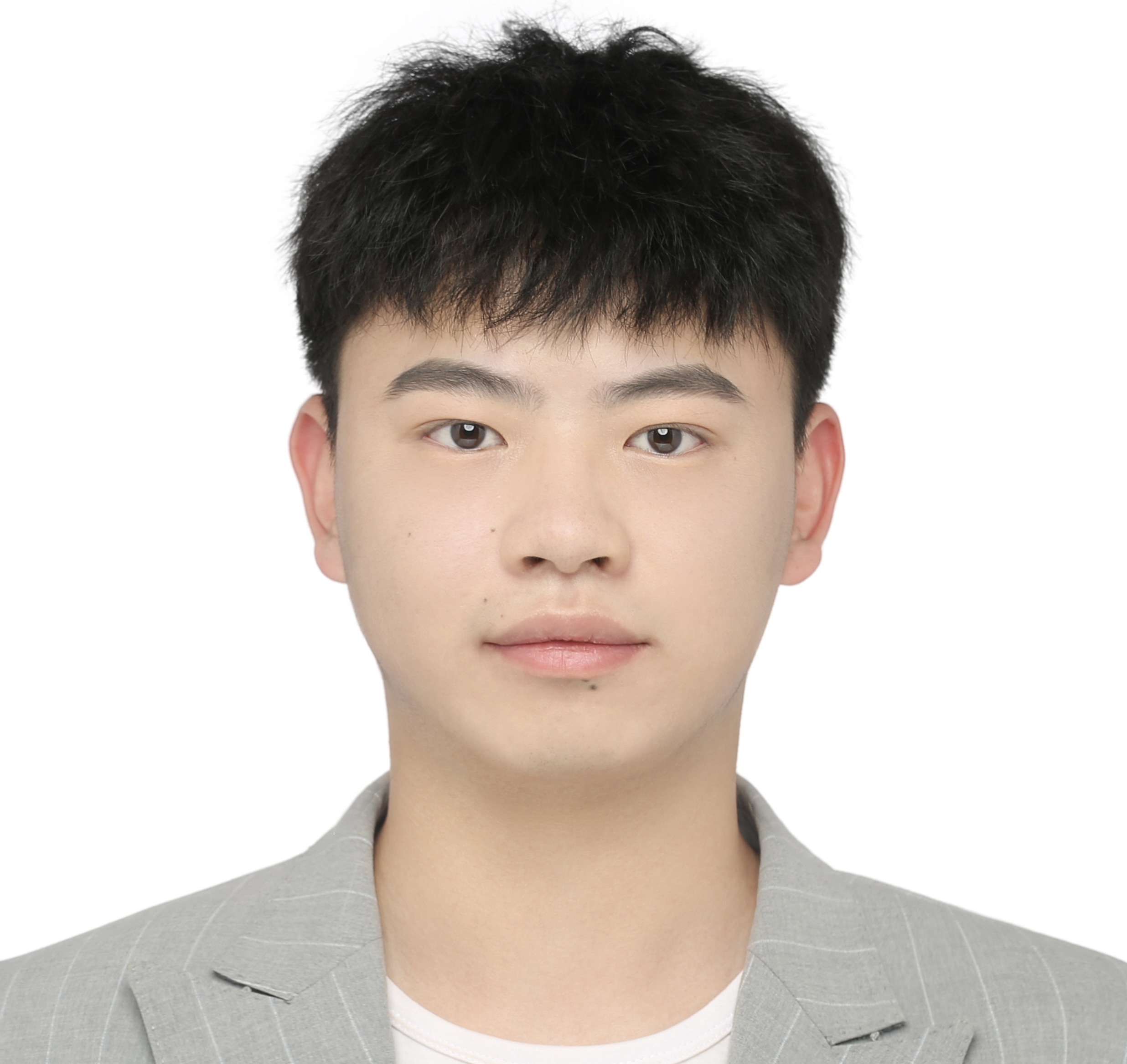}}]{Lin Long} received his B.Sc. degree in Computer Science and Technology from the College of Computer Science, Sichuan University, Chengdu, China, in 2025. He is currently a M.Sc. student in Electronic Information at the same institution. His current research interests include Agentic AI and multi-modal learning.
\end{IEEEbiography}
\vspace{-33pt}
\begin{IEEEbiography}[{\includegraphics[width=1in,height=1.25in,clip,keepaspectratio]{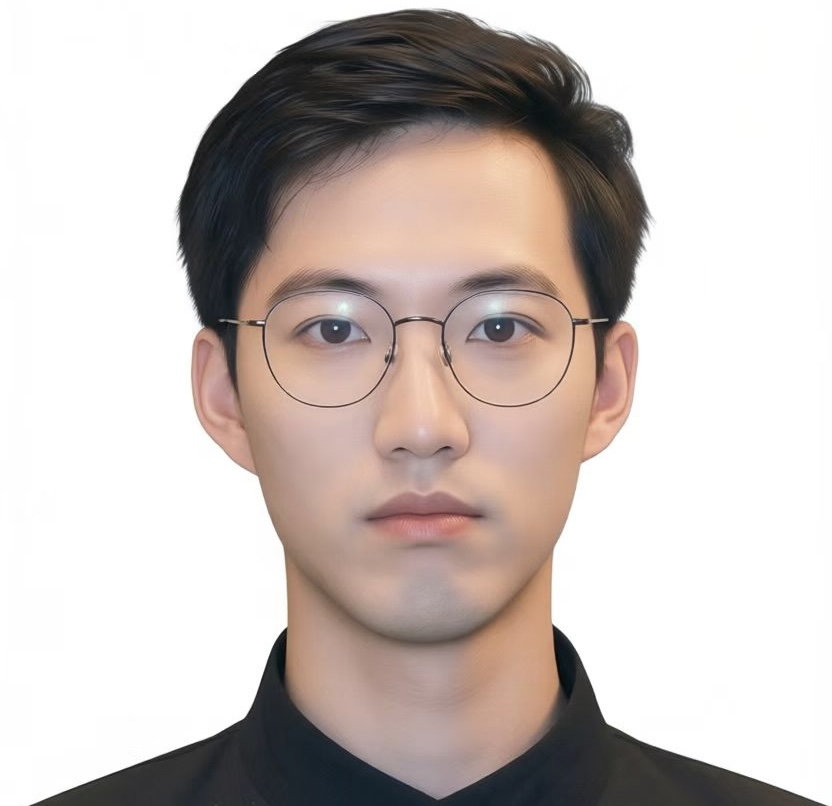}}]{Xinyu Liu} received his M.S. degree in Artificial Intelligence from the College of Computer Science, Sichuan University, in 2024. He is currently a Ph.D. student in Computer Science and Technology at the same institution. His current research interests include Agentic AI, AI safety, and vision-language action.
\end{IEEEbiography}
\vspace{-33pt}
\begin{IEEEbiography}[{\includegraphics[width=1in,height=1.25in,clip,keepaspectratio]{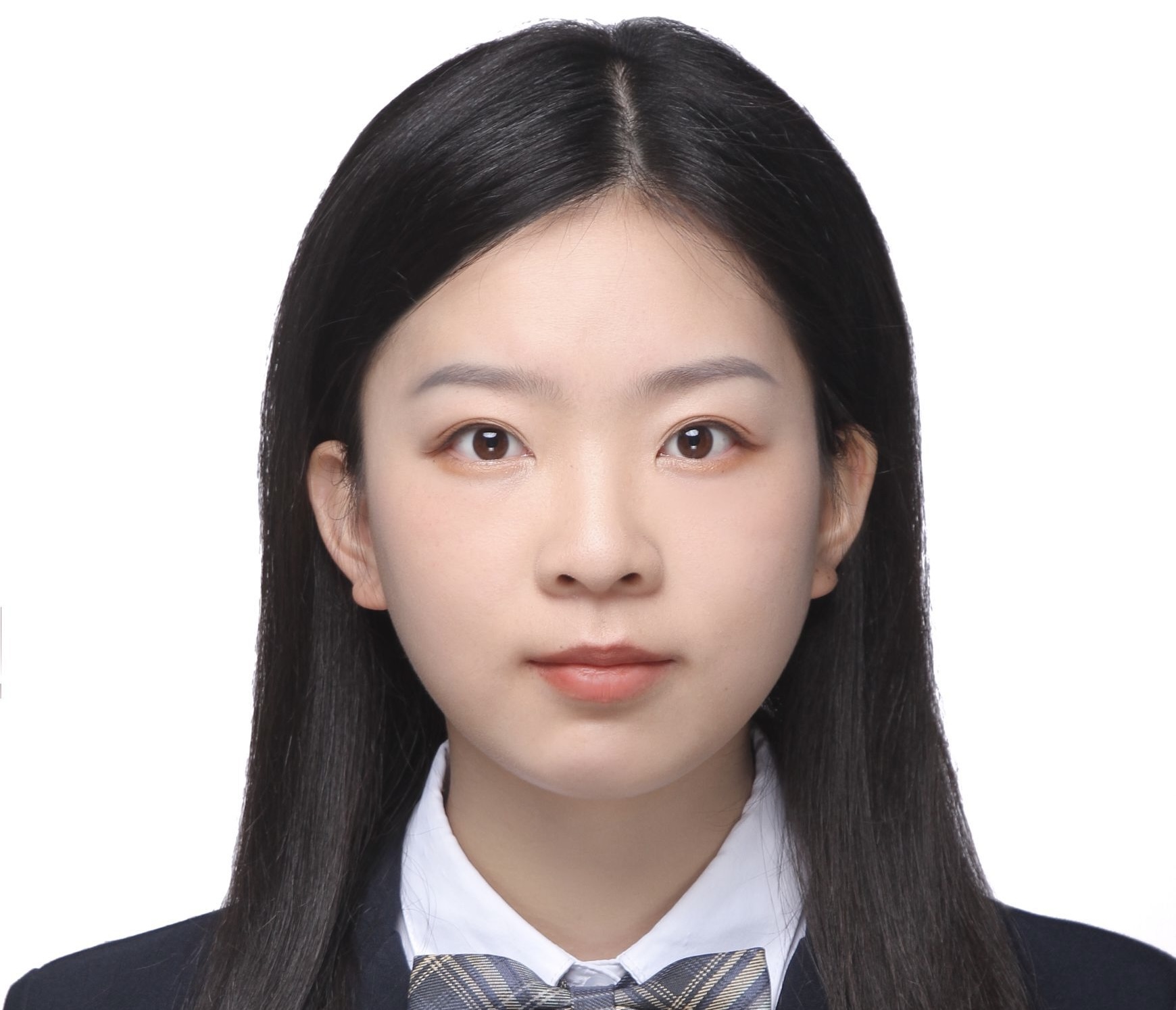}}]{Jingyu Xing} is currently pursuing the M.S. degree in Computer Science and Technology at the College of Computer Science, Sichuan University. Her research interests mainly focus on multi-modal learning, computer vision, and test-time adaption.
\end{IEEEbiography}
\vspace{-33pt}
\begin{IEEEbiography}[{\includegraphics[width=1in,height=1.25in,clip,keepaspectratio]{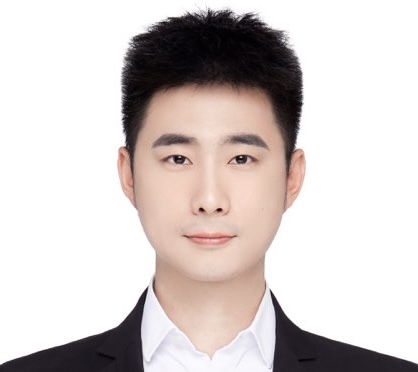}}]{Zizhou Wang} is a scientist with the Institute of High Performance Computing (IHPC), Agency for Science, Technology and Research (A*STAR), Singapore. He received the Ph.D. degree in computer science from Sichuan University, Chengdu, China, in 2022. His research interests include AI for healthcare, AI safety, and domain generalization.
\end{IEEEbiography}
\vspace{-33pt}
\begin{IEEEbiography}[{\includegraphics[width=1in,height=1.25in,clip,keepaspectratio]{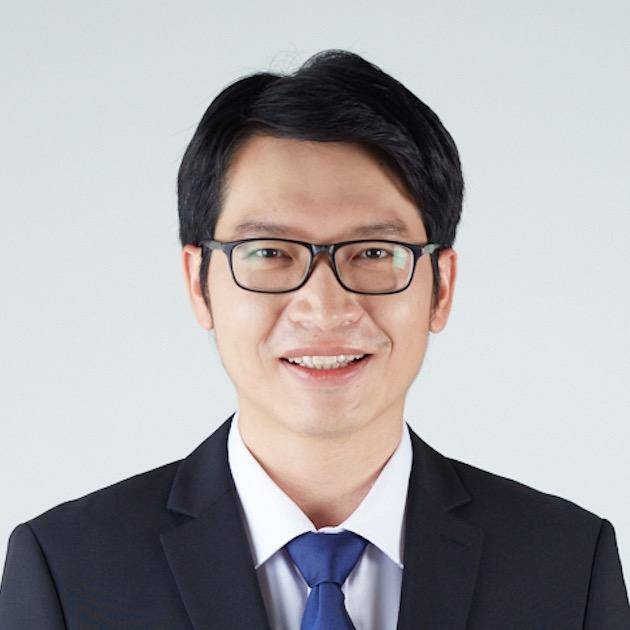}}]{Joey Tianyi Zhou} is the Deputy Director and Principal Investigator at the A*STAR Centre for Frontier AI Research (CFAR) and the Centre for Advanced Technologies in Online Safety (CATOS), Singapore. He received his Ph.D. from Nanyang Technological University (NTU) in 2015. Dr Zhou was recognized as one of the World’s Top 2\% Scientists (Career Lifetime) and was a finalist for the National Research Foundation Fellowship (NRF-F) in 2020.
\end{IEEEbiography}
\vspace{-33pt}
\begin{IEEEbiography}[{\includegraphics[width=1in,height=1.25in,clip,keepaspectratio]{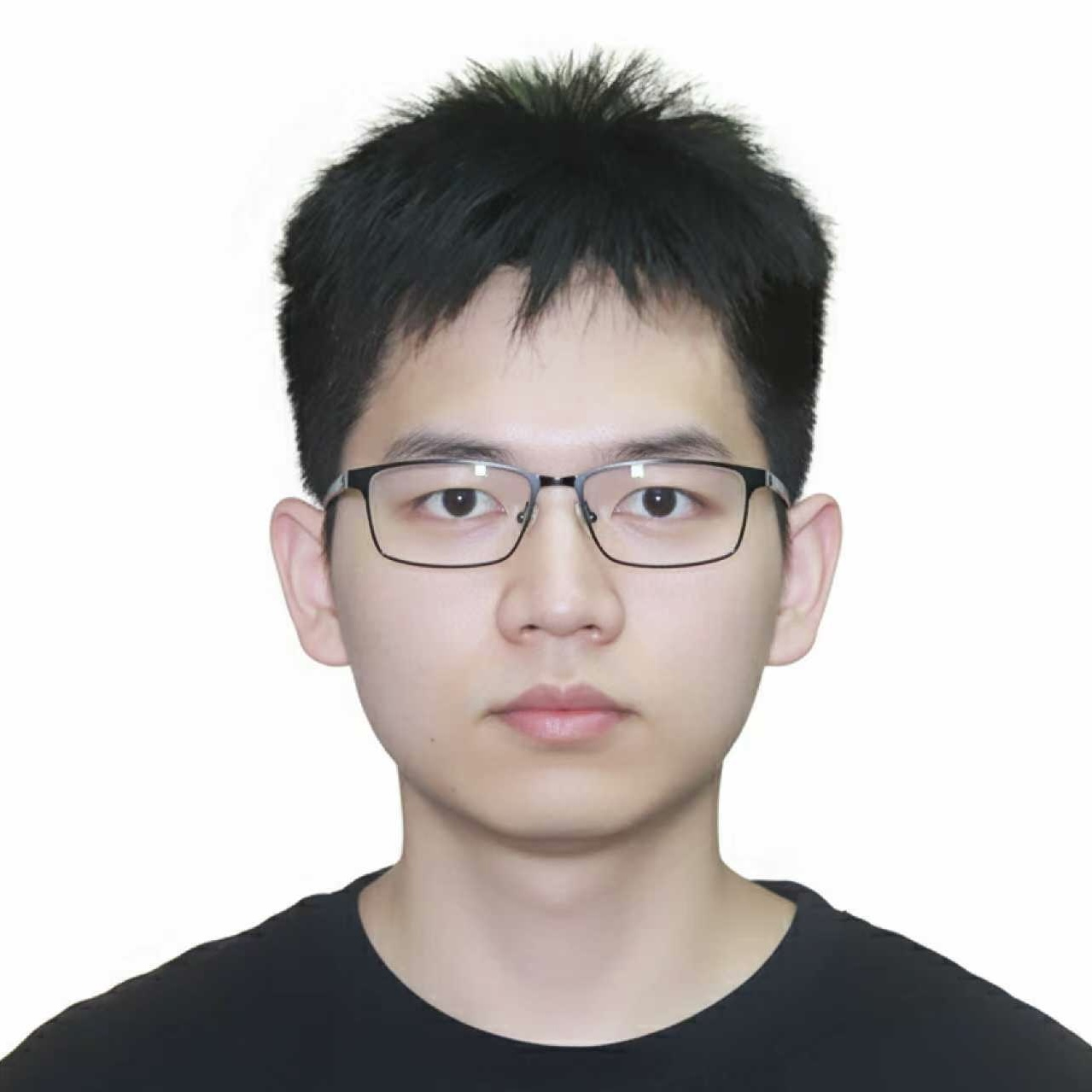}}]{Jiawei Du} received the Ph.D. degree in computer science from the National University of Singapore, Singapore, in 2023. He is currently a Senior Research Scientist with the Center for Frontier AI Research (CFAR) under the Agency for Science, Technology and Research, Singapore. His research primarily focuses on efficient optimization techniques in machine learning.
\end{IEEEbiography}
\vspace{-33pt}
\begin{IEEEbiography}[{\includegraphics[width=1in,height=1.25in,clip,keepaspectratio]{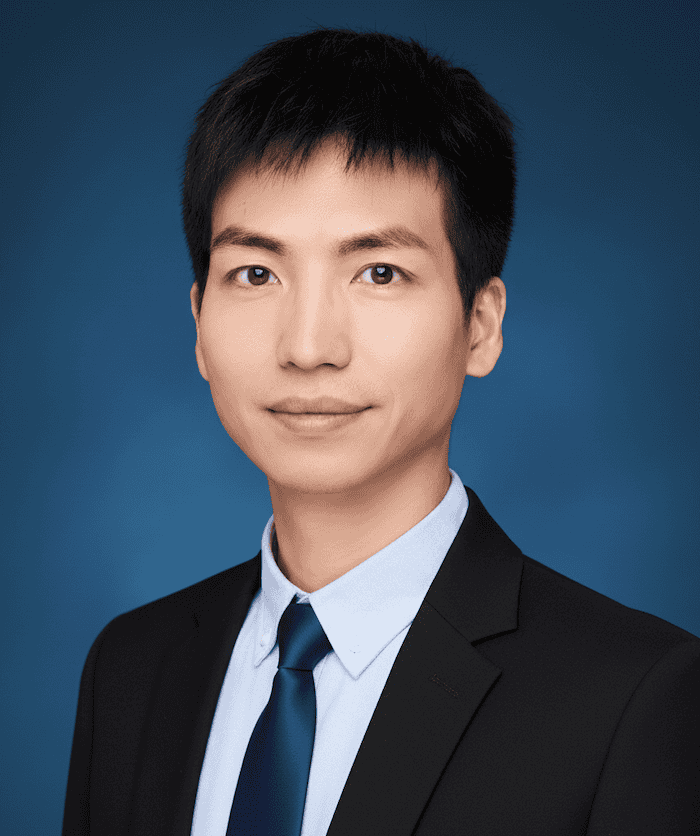}}]{Liangli Zhen} received the PhD degree from Sichuan University, in 2018. He is a senior scientist and group manager with the Institute of High Performance Computing (IHPC), Agency for Science, Technology and Research (A*STAR), Singapore. His research interests include machine learning and optimisation. He has led/co-led multiple projects under major national initiatives, including Singapore Aerospace Programme and AI Singapore Robust AI Grand Challenge.  
\end{IEEEbiography}
\vspace{-33pt}
\begin{IEEEbiography}[{\includegraphics[width=1in,height=1.25in,clip,keepaspectratio]{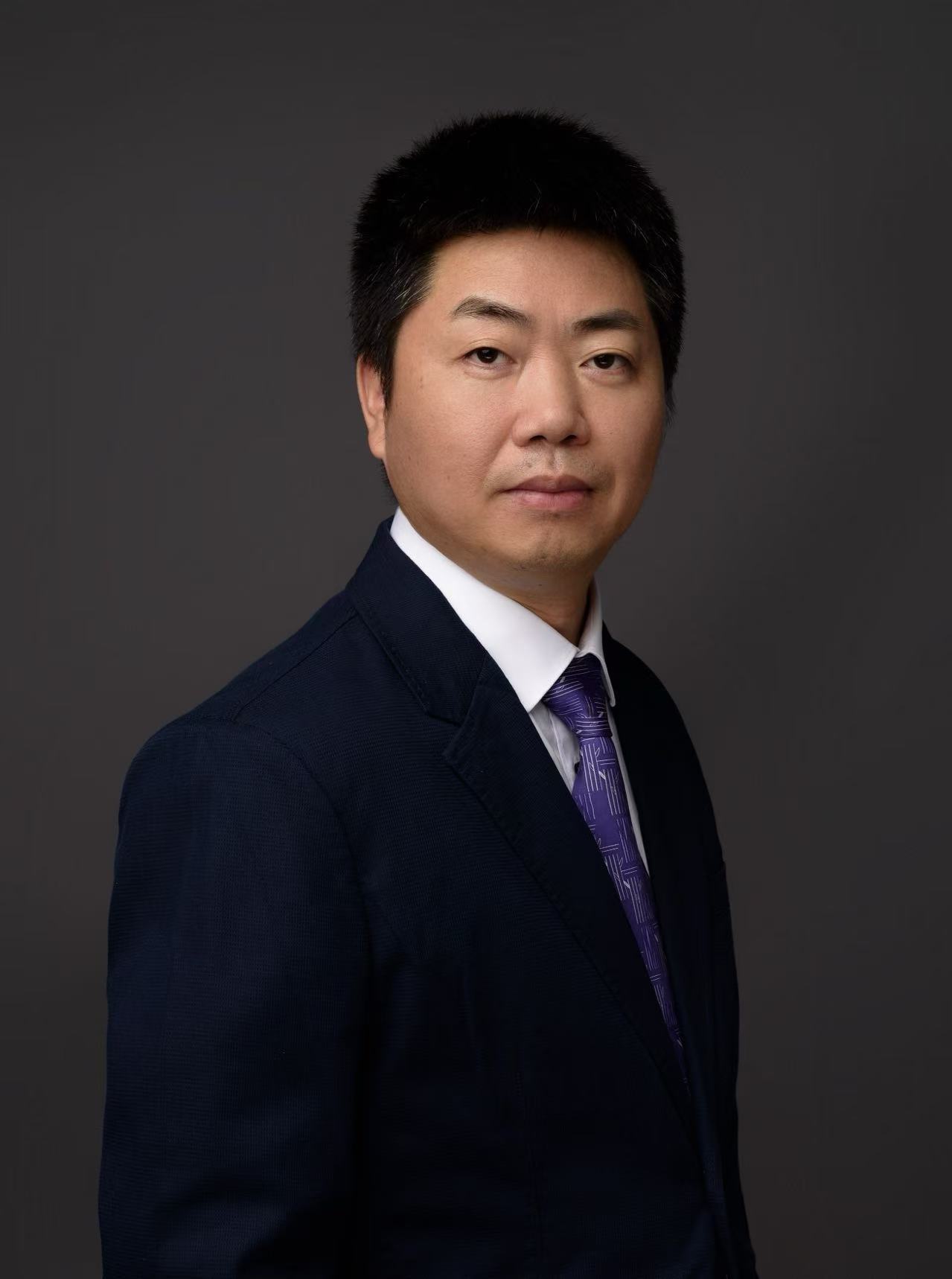}}]{Jiancheng Lv} received the Ph.D. degree in computer science and engineering from the University of Electronic Science and Technology of China, Chengdu, China, in 2006. He is currently a professor with the Data Intelligence and Computing Art Laboratory, College of Computer Science, Sichuan University, Chengdu, China. His current research interests include neural networks, deep learning, multi-modal learning, and artificial intelligence application.
\end{IEEEbiography}


\vfill

\end{document}